\documentclass[10pt,twocolumn,letterpaper]{article}

\usepackage{cvpr}
\usepackage{times}
\usepackage{epsfig}
\usepackage{graphicx}
\usepackage{amsmath}
\usepackage{amssymb}
\usepackage{enumitem}
\usepackage{textcomp}
\usepackage{subcaption}
\newcommand{\HZ}[1]{{\color{blue}{[Henry: #1]}}}

\newcommand{\Name}{GameGAN\ }

\usepackage[pagebackref=true,breaklinks=true,letterpaper=true,colorlinks,bookmarks=false]{hyperref}

\cvprfinalcopy 

\def\cvprPaperID{9714} 

\ifcvprfinal\fi
\begin{document}

\title{\vspace{-0mm}Learning to Simulate Dynamic Environments with  \Name}

\author{\vspace{1mm}Seung Wook Kim$^{1,2,3}\thanks{Correspondence to {\tt\small \{seungwookk,sfidler\}@nvidia.com}}$
\hspace{0.5cm}
Yuhao Zhou$^{2}$\thanks{YZ worked on this project during his internship at NVIDIA.}
\hspace{0.5cm}
Jonah Philion$^{1,2,3}$
\hspace{0.5cm}
Antonio Torralba$^{4}$
\hspace{0.5cm}
Sanja Fidler$^{1,2,3}$\footnotemark[1]
\\
$^1$NVIDIA 
\hspace{2em} 
$^2$University of Toronto 
\hspace{2em} 
$^3$Vector Institute  
\hspace{2em} 
$^4$ MIT
\\
{\tt\small \{seungwookk,jphilion,sfidler\}@nvidia.com}
\hspace{2em}
{\tt\small henryzhou@cs.toronto.edu}
\hspace{2em}
{\tt\small torralba@mit.edu}
\\
\url{https://nv-tlabs.github.io/gameGAN}
}

\maketitle

\begin{abstract}

Simulation is a crucial component of any robotic system. In order to simulate correctly, we need to write complex rules of the environment: how dynamic agents behave, and how the actions of each of the agents affect the behavior of others. In this paper, we aim to \emph{learn} a simulator by simply watching an agent interact with an environment.
We focus on graphics games  as a proxy of the real environment. 
We introduce GameGAN, a generative model that learns to visually imitate a desired game by ingesting screenplay and keyboard actions during training. Given a key pressed by the agent, \Name ``renders" the next screen using a carefully designed generative adversarial network. Our approach offers key advantages over existing work: 
we design a memory module that builds an internal map of the environment, allowing for the agent to return to previously visited locations with high visual consistency. 
In addition, \Name is able to disentangle static and dynamic components within an image making the behavior of the model more interpretable, and relevant for downstream tasks that require explicit reasoning over dynamic elements.
This enables  many interesting applications such as swapping different components of the game to build new games that do not exist.
\end{abstract}

\vspace{-3mm}
\section{Introduction}

Before deployment to the real world, an artificial agent needs to undergo extensive testing in challenging simulated environments. Designing good simulators is thus extremely important. This is traditionally done by writing procedural models to generate valid and diverse scenes, and complex behavior trees that specify how each actor in the scene behaves and reacts to actions made by other actors, including the ego agent. 
However, writing simulators that encompass a large number of diverse scenarios is extremely time consuming and requires highly skilled graphics experts. 
Learning to simulate by simply observing the dynamics of the real world is the most scaleable way going forward.

\begin{figure}[t!]
\vspace{-1mm}
\begin{center}
\includegraphics[width=8.0cm]{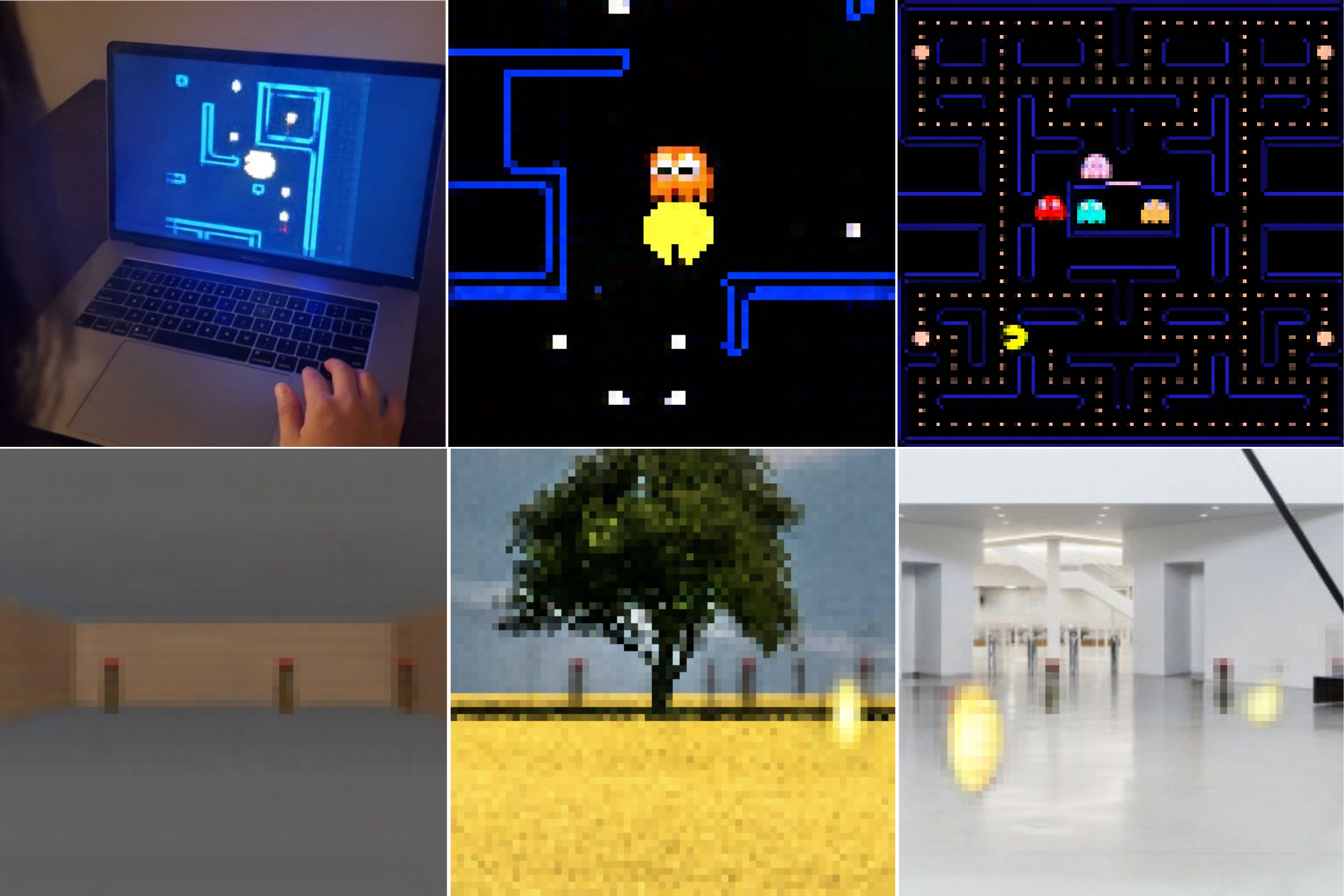}
\end{center}
\vspace{-5mm}
\caption{\footnotesize If you look at the person on the top-left picture, you might think she is playing Pacman of Toru Iwatani, but she is not!
      She is actually playing with a GAN generated version of Pacman.
      In this paper, we introduce \Name that  learns to reproduce games by just observing lots of playing rounds.
      Moreover, our model can disentangle background from dynamic objects, allowing us to create new games by swapping components  as shown in the center and right images of the bottom row.}
\label{fig:teaser}
\vspace{-5mm}
\end{figure}

\begin{figure*}[!ht]
\vspace{-3mm}
\begin{center}
\includegraphics[width=0.8\textwidth]{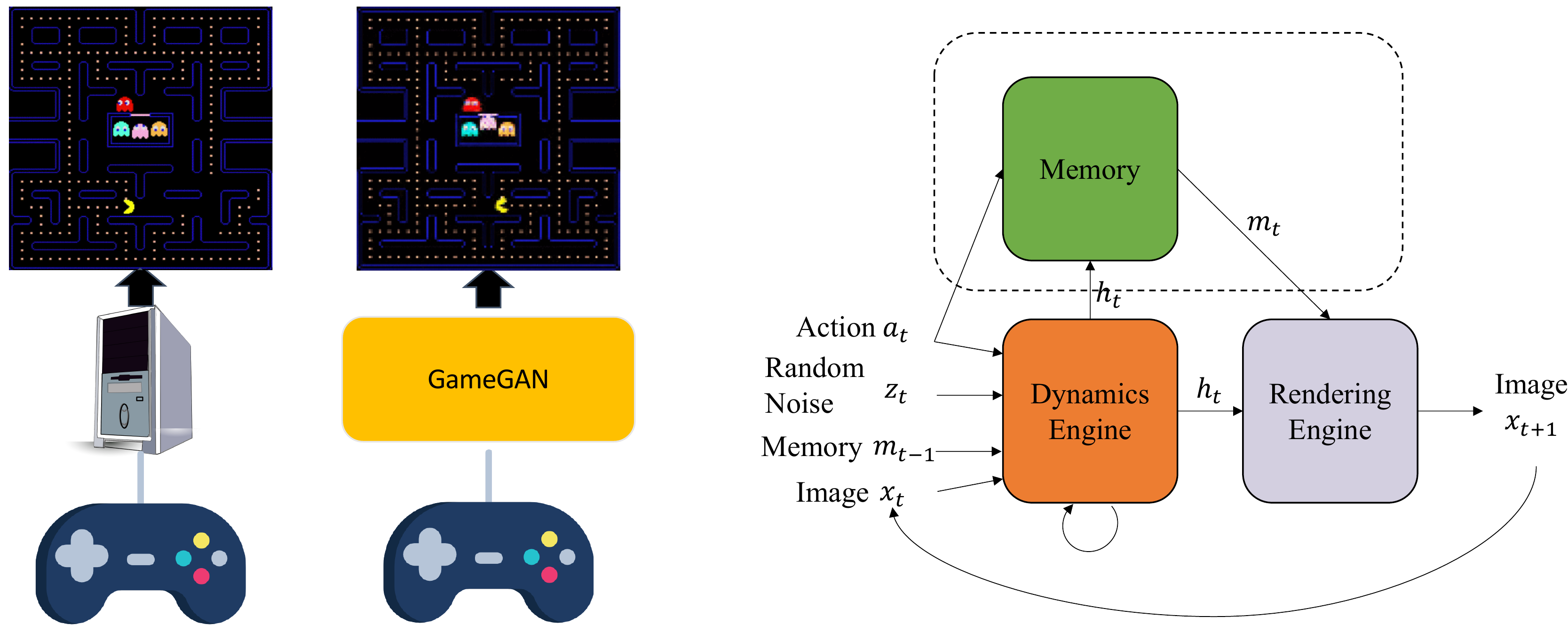}
\end{center}
\vspace{-4mm}
\caption{\footnotesize Overview of GameGAN: Our goal is to replace the game engine with neural networks. \Name is composed of three main modules. The \emph{dynamics engine} is implemented as an RNN, and contains the world state that is updated at each time $t$.
        Optionally, it can write to and read from the external \emph{memory module} $\mathcal{M}$.
        Finally, the \emph{rendering engine} is used to decode the output image.
        All modules are neural networks and trained end-to-end.}
\label{fig:overview}
\vspace{-3mm}
\end{figure*}

A plethora of existing work aims at learning behavior models~\cite{beer1992evolving,paxton2017combining,jain2019discrete,chen2017learning}. However, these typically assume  a significant amount of supervision such as access to agents' ground-truth trajectories. 
 We aim to learn a simulator by simply watching an agent interact with an environment. To simplify the problem, we frame this as a 2D image generation problem. Given sequences of observed image frames and the corresponding  actions the agent took, we wish to emulate image creation as if ``rendered" from a real dynamic environment that is reacting to the agent's actions. 

Towards this goal, we focus on graphics games, which represent simpler and more controlled environments, as a proxy of the real environment. Our goal is to replace the graphics engine at test time, by visually imitating the game using a learned model. 
This is a challenging problem:  different games have different number of components as well as different physical dynamics. 
Furthermore, many games require long-term consistency in the environment.
For example, imagine a game where an agent navigates through a maze.
When the agent moves away and later returns to a location, it expects the scene to look consistent with what it has encountered before. 
In visual SLAM, \emph{detecting} loop closure (returning to a previous location) is already known to be challenging, let alone \emph{generating} one. 
Last but not least, both deterministic and stochastic behaviors typically exist in a game, and modeling the latter is known to be particularly hard.


In this paper, we introduce GameGAN, a generative model that learns to imitate a desired game. GameGAN  ingests screenplay and keyboard actions during training and aims to predict the next frame by conditioning on the action, \ie a key pressed by the agent. 
It learns from rollouts of image and action pairs directly without having access to the underlying game logic or engine. We make several advancements over the recently introduced World Model~\cite{ha2018recurrent} that aims to solve a similar problem. By leveraging Generative Adversarial Networks ~\cite{goodfellow2014generative}, we produce higher quality results. Moreover, while~\cite{ha2018recurrent} employs a straightforward conditional decoder, \Name features a carefully designed architecture. In particular, we propose a new memory module that encourages the model to build an internal map of the environment, allowing the agent to return to previously visited locations with high visual consistency. Furthermore, we introduce a purposely designed decoder that learns to disentangle static and dynamic components within the image.  This makes the behavior of the model more interpretable, and it further allows us to modify  existing games by swapping out different components.

We test \Name on a modified version of Pacman and the VizDoom environment~\cite{kempka2016vizdoom}, and propose several synthetic tasks for both quantitative and qualitative evaluation.
We further introduce a come-back-home task to test the long-term consistency of learned simulators. 
Note that \Name supports several applications such as transferring a given game from one operating system to the other, without requiring to re-write code.

\begin{figure*}[!h]
    \vspace{-2mm}
    \begin{center}
        \includegraphics[width=\textwidth]{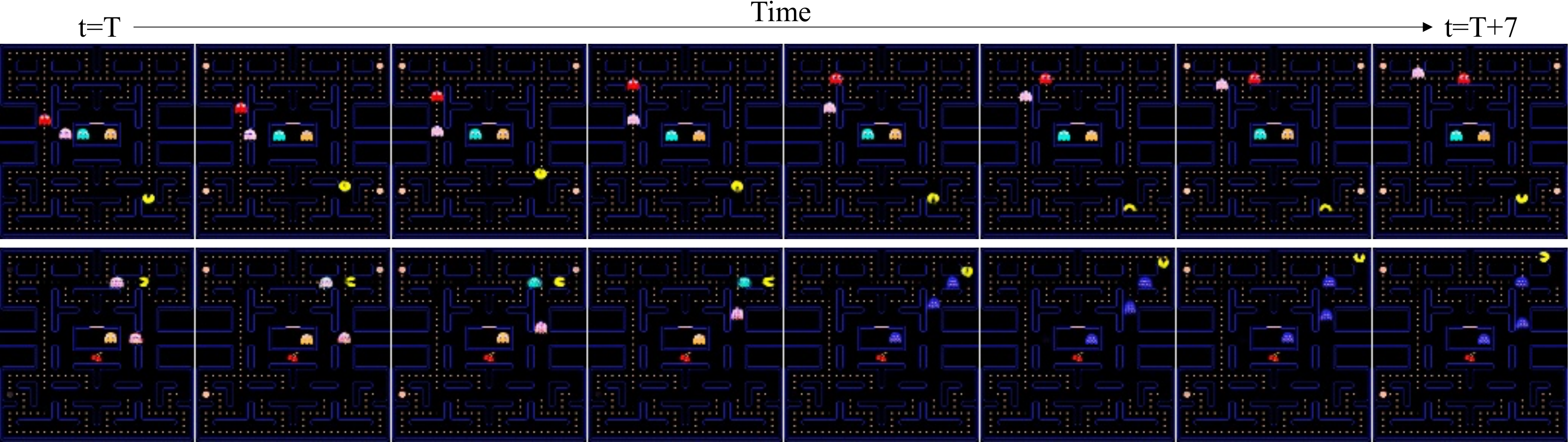}
    \end{center}
    \vspace{-3mm}
    \caption[Caption for LOF]{
        Screenshots of a human playing with \Name trained on the official version of Pac-Man\protect\footnotemark.
        \Name learns to produce a visually consistent simulation as well as learning the dynamics of the game well.
        On the bottom row, the player consumes a capsule, turning the ghosts purple.
        Note that ghosts approach Pacman before consuming the capsule, and run away after.
        
    }
    \label{fig:pacman_bandai_rollout}
    \vspace{-2mm}
\end{figure*}

\vspace{-0mm}
\section{Related Work}

\vspace{-0mm}
\paragraph{Generative Adversarial Networks:}
 In GANs~\cite{goodfellow2014generative}, a generator and a discriminator play an adverserial game that encourages the generator to produce realistic outputs. 
 To obtain a desired control over the generated outputs, categorical labels~\cite{miyato2018cgans}, images~\cite{pix2pix_isola2016, img2img_translation_liu_2017}, captions~\cite{DBLP:conf/icml/ReedAYLSL16}, or masks~\cite{park2019semantic} are provided as input to the generator.
 Works such as~\cite{vid2vid2018} synthesize new videos by transferring the style of the source to the target video using the cycle consistency loss~\cite{CycleGAN2017,kim2017learning}. Note that this is a simpler problem than the problem considered in our work, as the dynamic content of the target video is provided and only the visual style needs to be modified.
 In this paper, we consider generating the dynamic content itself. We adopt the GAN framework and use the user-provided action as a condition for generating future frames.
 To the best of our knowledge, ours is the first work on using action-conditioned GANs  for emulating game simulators.

\vspace{-5mm}
\paragraph{Video Prediction:}
 Our work shares similarity to the task of video prediction  which aims at predicting future frames given a sequence of previous frames.
 Several works~\cite{srivastava2015unsupervised, chiappa2017recurrent, NIPS2015_5859} train a recurrent encoder to decode future frames.   Most approaches are trained with a reconstruction loss, resulting in a deterministic process that generates blurry frames and often does not handle stochastic behaviors well. The errors typically accumulate over time and result in low quality predictions.
 Action-LSTM models~\cite{chiappa2017recurrent, NIPS2015_5859} achieved success in scaling the generated images to higher resolution but do not handle complex stochasticity present in environments like Pacman.
 Recently,~\cite{ha2018recurrent, denton2018stochastic} proposed VAE-based frameworks to capture the stochasticity of the task. However, the resulting videos are blurry and the generated frames tend to omit certain details.
GAN loss has been previously used in several works \cite{denton_dr_NIPS2017_7028,DBLP:journals/corr/abs-1804-01523,tulyakov2018mocogan,clark2019efficient}.
 \cite{denton_dr_NIPS2017_7028} uses an adversarial loss to disentangle pose from content across different videos. 
 In \cite{DBLP:journals/corr/abs-1804-01523}, VAE-GAN \cite{larsen2015autoencoding} formulation is used for generating the next frame of the video.
Our model differs from these works in that in addition to generating the next frame, \Name also learns the intrinsic dynamics of the environment.
\cite{guzdial2017game} learns a game engine by parsing frames and learning a rule-based search engine that finds the most likely next frame. \Name goes one step further by learning everything from frames and also learning to directly generate frames.

\vspace{-4mm}
\paragraph{World Models:}
 In model-based reinforcement learning, one uses interaction with the environment to learn a dynamics model.
 World Models~\cite{ha2018recurrent} exploit a learned simulated environment to train an RL agent instead.
 Recently, World Models have been used to generate Atari games in a concurrent work~\cite{anonymous2020model}.
 The key differences with respect to these models are in the design of the architecture: we introduce a memory module to better capture long-term consistency, and a carefully designed decoder that disentangles static and dynamic components of the game.


\vspace{-1mm}
\section{\Name}
\label{sec:model}
\vspace{-0mm}

\footnotetext{PAC-MAN\texttrademark \& \textcopyright BANDAI NAMCO Entertainment Inc.}

We are interested in training a game simulator that can model both deterministic and stochastic nature of the environment.
In particular, we focus on an action-conditioned simulator in the image space where there is an egocentric agent that moves according to the given action $a_t \sim \displaystyle \mathcal{A}$ at time $t$ and generates a new observation $x_{t+1}$.
We assume there is also a stochastic variable $z_t \sim \displaystyle \mathcal{N} (0; I)$ that corresponds to randomness in the environment.
Given the history of images $x_{1:t}$ along with $a_t$ and $z_t$, \Name predicts the next image $x_{t+1}$.
\Name is composed of three main modules. 
The \emph{dynamics engine} (Sec~\ref{sec:dynamics_engine}), which maintains an internal state variable, takes $a_t$ and $z_t$ as inputs and updates the current state.
For environments that require long-term consistency, we can optionally use an external \emph{memory module} (Sec~\ref{sec:memory}).
Finally, the \emph{rendering engine} (Sec~\ref{sec:rendering_engine}) produces the output image given the state of the dynamics engine.
It can be implemented as a simple convolutional decoder or can be coupled with the memory module to disentangle static and dynamic elements while ensuring long-term consistency.
We use adversarial losses along with a proposed temporal cycle loss (Sec~\ref{sec:losses}) to train GameGAN.
Unlike some works ~\cite{ha2018recurrent} that use sequential training for stability, \Name is trained end-to-end.
We provide more details of each module in the supplementary materials.

\vspace{-1mm}
\subsection{Dynamics Engine}
\label{sec:dynamics_engine}
\vspace{-1mm}
\Name has to learn how various aspects of an environment change with respect to the given user action.
For instance, it needs to learn that certain actions are not possible (\eg walking through a wall), and how other objects behave as a consequence of the action.
We call the primary component that learns such transitions the \emph{dynamics engine} (see illustration in Figure~\ref{fig:overview}).
It needs to have access to the past history to produce a consistent simulation.
Therefore, we choose to implement it as an action-conditioned LSTM \cite{hochreiter_lstm1997}, motivated by the design of Chiappa \etal\cite{chiappa2017recurrent}:

\begin{equation}
v_t = h_{t-1} \odot \mathcal{H}(a_t, z_t, m_{t-1}),  s_t = \mathcal{C}(x_t)
\end{equation}
\begin{equation}
    \begin{gathered}
        i_t = \sigma(W^{iv} v_t + W^{is} s_t),
        f_t = \sigma(W^{fv} v_t + W^{fs} s_t), \\
        o_t = \sigma(W^{ov} v_t + W^{os} s_t)
    \end{gathered}
\end{equation}
\begin{equation}
c_t = f_t \odot c_{t-1} + i_t \odot \tanh(W^{cv}v_t + W^{cs} s_t)
\end{equation}
\begin{equation}
h_t = o_t \odot \tanh(c_t)
\end{equation}
where $h_t, a_t, z_t, c_t, x_t$ are the hidden state, action, stochastic variable, cell state, image at time step $t$.
$m_{t-1}$ is the retrieved memory vector in the previous step (if the memory module is used), and $i_t, f_t, o_t$ are the input, forget, and output gates.
$a_t$, $z_t, m_{t-1}$ and $h_t$ are fused into $v_t$, and $s_t$ is the encoding of the image $x_t$.
$\mathcal{H}$ is a MLP, $\mathcal{C}$ is a convolutional encoder, and $W$ are weight matrices.
$\odot$ denotes the hadamard product.
The engine maintains the standard state variables for LSTM, $h_t$ and $c_t$, which contain information about every aspect of the current environment at time $t$.
It computes the state variables given $a_t$, $z_t$, $m_{t-1}$, and  $x_t$.

\vspace{-1mm}
\subsection{Memory Module}
\label{sec:memory}
\vspace{-1mm}
\begin{figure}
\vspace{-3mm}
\begin{center}
\includegraphics[width=\linewidth]{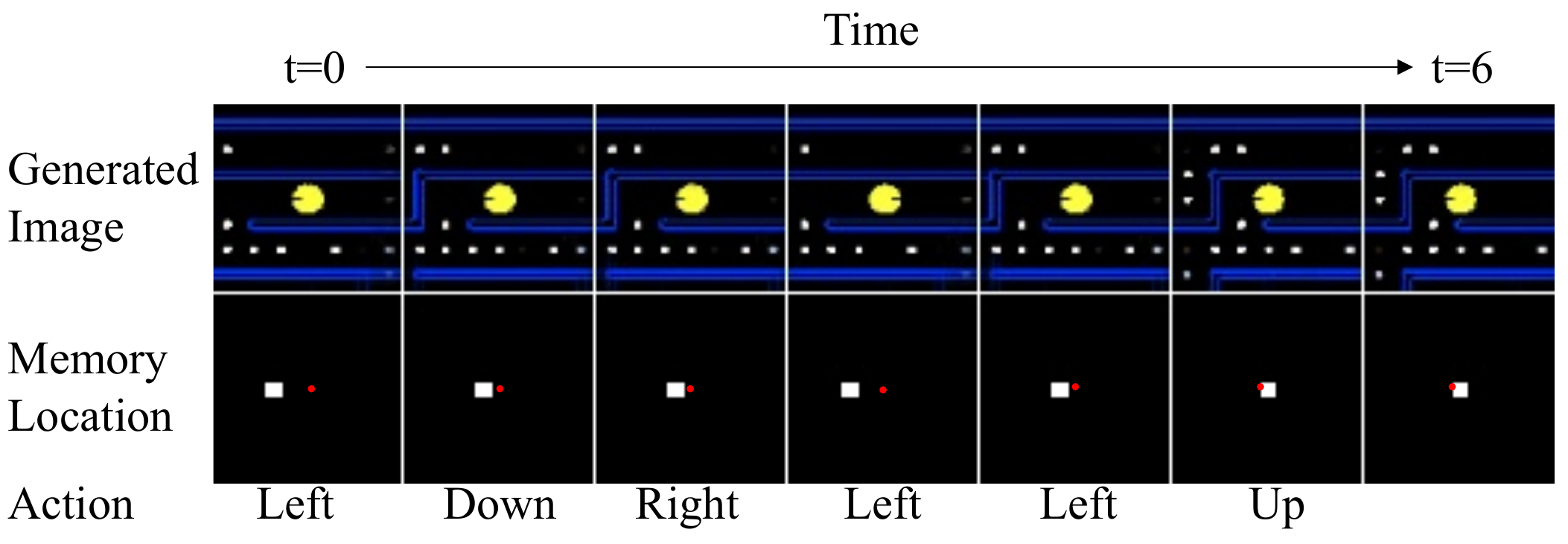}
\end{center}
\vspace{-4mm}
\caption{\footnotesize Visualizing attended memory location $\alpha$:
        Red dots marking the center are placed to aid visualization.
        Note that we \emph{learn} the memory shift, so the user action does not always align with how the memory is shifted.
         In this case, \emph{Right} shifts $\alpha$ to the left, and \emph{Left} shifts $\alpha$ to the right.
         It also \emph{learns} not to shift when an invalid action is given.
         }
\label{fig:memory_rollout}
\vspace{-6mm}
\end{figure}

Suppose we are interested in simulating an environment in which there is an agent navigating through it.
This requires long-term consistency in which the simulated scene (\eg buildings, streets) should not change when the agent comes back to the same location a few moments later.
This is a challenging task for typical models such as RNNs because 1) the model needs to remember every scene it generates in the hidden state, and 2) it is non-trivial to design a loss that enforces such long-term consistency.
We propose to use an external memory module, motivated by the Neural Turing Machine (NTM) \cite{graves2014neural}. 

The memory module has a memory block $\mathcal{M} \in \mathbb{R}^{N\times N\times D}$, and the attended location $\alpha_t \in \mathbb{R}^{N\times N}$ at time $t$.
$\mathcal{M}$ contains $N\times N$ $D$-dimensional vectors where $N$ is the spatial width and height of the block. 
Intuitively, $\alpha_t$ is the current location that the egocentric agent is located at.
$\mathcal{M}$ is initialized with random noise $\sim N(0,I)$ and $\alpha_0$ is initialized with 0s except for the center location $(N/2, N/2)$ that is set to 1.
At each time step, the memory module computes:
\begin{equation}
w = \mathrm{softmax}(\mathcal{K}(a_t)) \in \mathbb{R}^{3\times 3}
\end{equation}
\begin{equation}
g = \mathcal{G}(h_t) \in \mathbb{R}
\end{equation}
\begin{equation}
\alpha_{t} = g \cdot \mathrm{Conv2D}(\alpha_{t-1}, w) + (1-g)\cdot \alpha_{t-1}
\end{equation}
\begin{equation}
\mathcal{M} = \mathrm{write}(\alpha_t, \mathcal{E}(h_t), \mathcal{M}) 
\end{equation}
\begin{equation}
m_t = \mathrm{read}(\alpha_t, \mathcal{M}) 
\end{equation}
where $\mathcal{K}, \mathcal{G}$ and $\mathcal{E}$ are small MLPs.
$w$ is a learned shift kernel that depends on the current action, and the kernel is used to shift $\alpha_{t-1}$.
In some cases, the shift should not happen (\eg cannot go forward at a dead end).
With the help from $h_t$, we also learn a gating variable $g \in [0,1]$ that determines if $\alpha$ should be shifted or not.
$\mathcal{E}$ is learned to extract information to be written from the hidden state.
Finally, $write$ and $read$ operations softly access the memory location specified by $\alpha$ similar to other neural memory modules \cite{graves2014neural}.
Using this shift-based memory module allows the model to not be bounded by the block $\mathcal{M}$'s size while enforcing local movements.
Therefore, we can use any arbitrarily sized block at test time.
Figure \ref{fig:memory_rollout} demonstrates the learned memory shift.
Since the model is free to assign actions to different kernels, the learned shift does not always correspond to how humans would do.
We can see that \emph{Right} is assigned as a left-shift, and hence \emph{Left} is assigned as a right-shift.
Using the gating variable $g$, it also learns not to shift when an invalid action, such as going through a wall, is given.

Enforcing long-term consistency in our case refers to remembering generated static elements (\eg background) and retrieving them appropriately when needed.
Accordingly, the benefit of using the memory module would come from storing static information inside it.
Along with a novel cycle loss (Section \ref{sec:cycle_loss}), we introduce inductive bias in the architecture of the rendering engine (Section \ref{sec:rendering_engine}) to encourage the disentanglement of static and dynamic elements.

\begin{figure}
    \vspace{-3mm}
        \begin{center}
        \includegraphics[width=8.5cm]{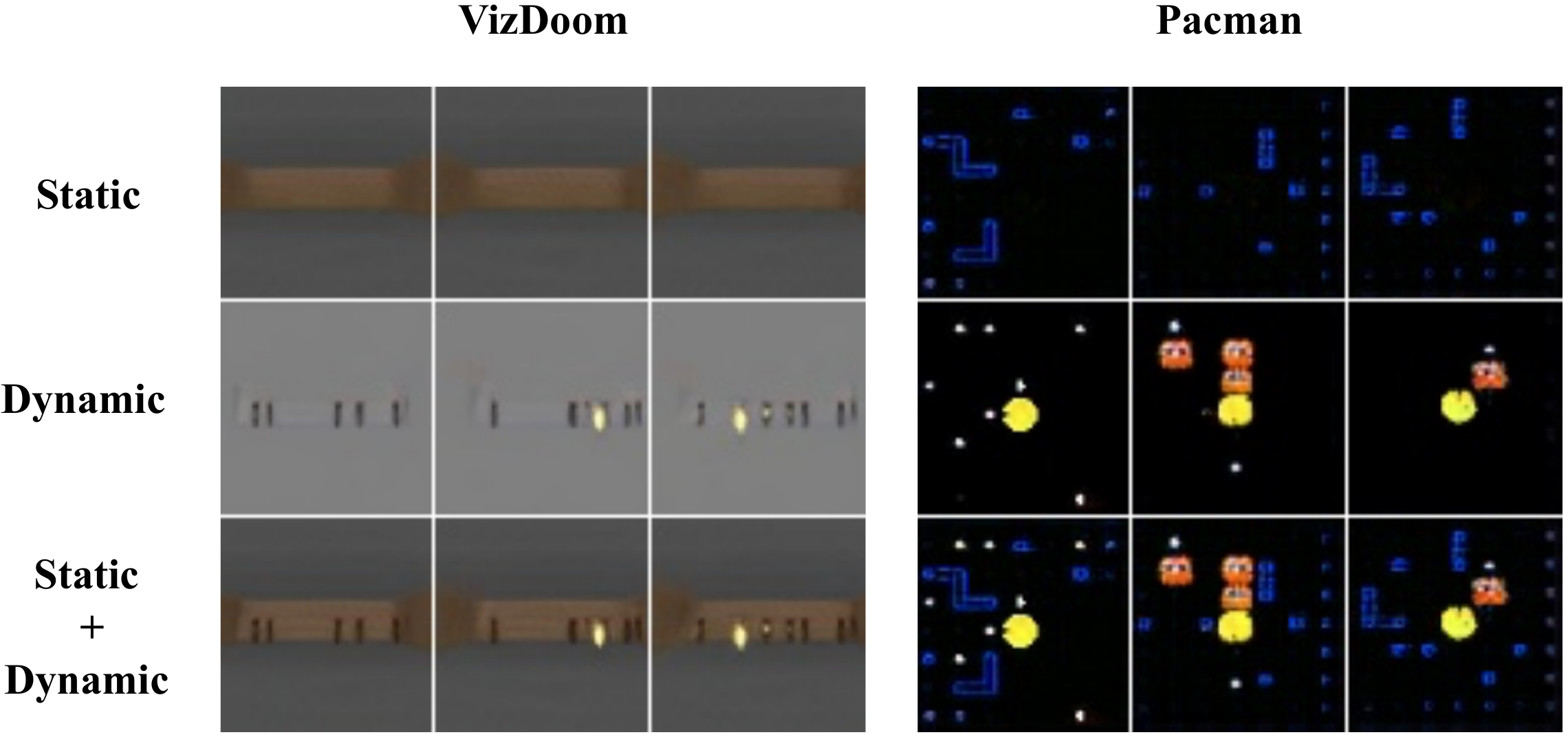}
        \end{center}
    \vspace{-4mm}
    \caption{
        \footnotesize Example showing how static and dynamic elements are disentangled in VizDoom and Pacman games with \Name.
        Static components usually include environmental objects such as walls.
        Dynamic elements typically are objects that can change as the game progresses such as food and other non-player characters.
    }
    \label{fig:disentangle_example}
    \vspace{-2mm}
\end{figure}

\subsection{Rendering Engine}
\label{sec:rendering_engine}
\begin{figure}
\begin{center}
\includegraphics[width=8.5cm]{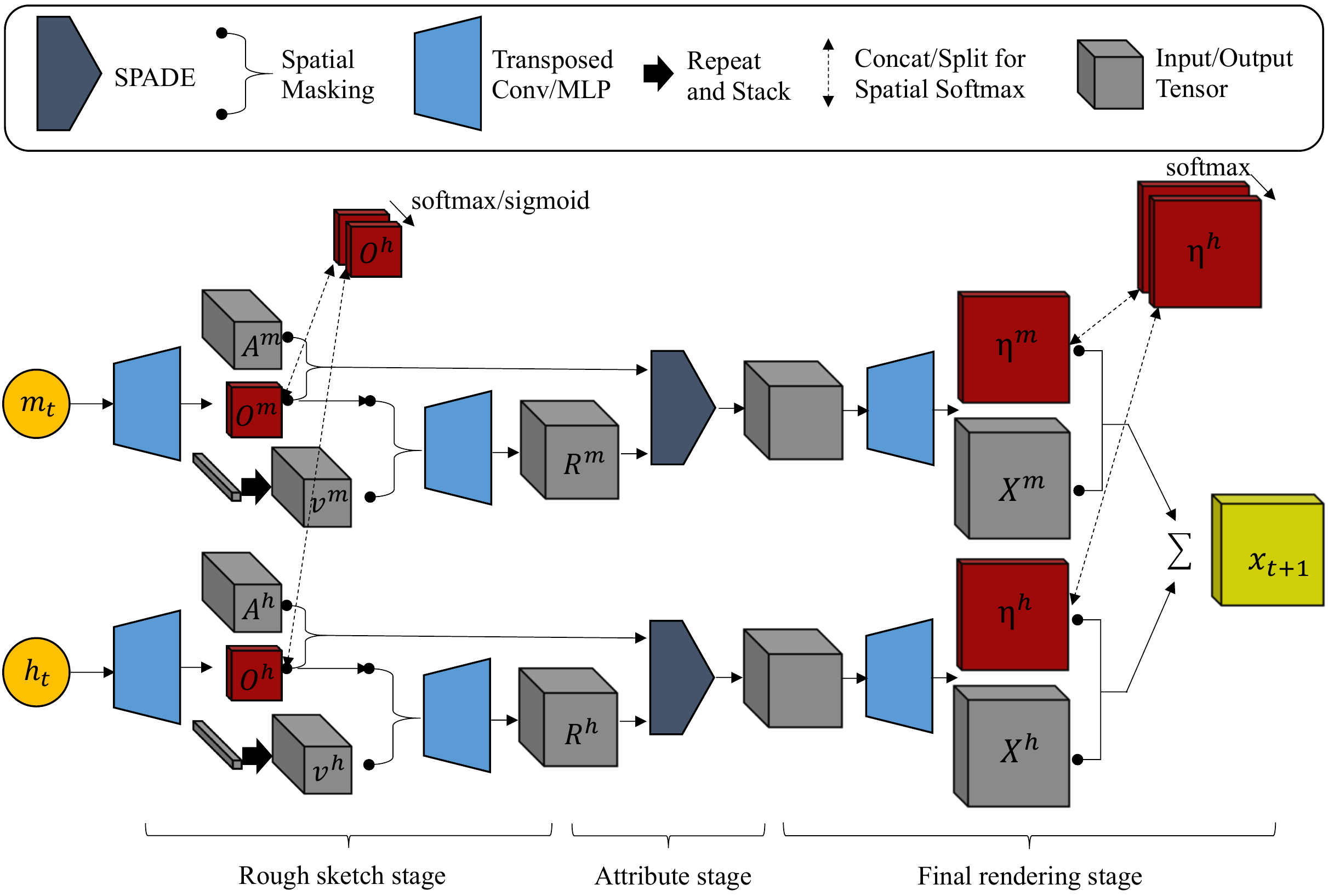}
\end{center}
\vspace{-3mm}
\caption{\footnotesize \emph{Rendering engine} for disentangling static and dynamic components. See Sec~\ref{sec:rendering_engine} for details. }
\label{fig:renderer}
\vspace{-4mm}
\end{figure}

The (neural) \emph{rendering engine} is responsible for rendering the simulated image $x_{t+1}$ given the state $h_t$.
It can be simply implemented with standard transposed convolution layers.
However, we also introduce a specialized rendering engine architecture (Figure \ref{fig:renderer}) for ensuring long-term consistency by learning to produce disentangled scenes.
In Section~\ref{sec:experiments}, we compare the benefits of each architecture.

The specialized rendering engine takes a list of vectors $\mathbf{c} = \{c^1,...,c^K\}$ as input.
In this work, we let $K=2$, and $\mathbf{c} = \{m_t, h_t\}$.
Each vector $c^k$ corresponds to one type of entity and goes through the following three stages (see Fig~\ref{fig:renderer}).
First, $c^{k}$ is fed into convolutional networks to produce an attribute map $A^k \in \displaystyle \mathbb{R}^{H_1\times H_1\times D_1}$ and object map $O^k \in \displaystyle \mathbb{R}^{H_1\times H_1\times 1}$.
It is also fed into a linear layer to get the type vector $v^k \in \displaystyle \mathbb{R}^{D_1}$ for the $k$-th component.  
$O$ for all components are concatenated together and fed through either a sigmoid to ensure $0 \leq O^k[x][y] \leq 1$ or a spatial softmax function so that $\sum^K_{k=1} O^k[x][y] = 1$ for all $x,y$.
The resulting object map is multiplied by the type vector $v^k$ in every location and fed into a convnet to produce $R^k \in \mathbb{R}^{H_2\times H_2\times D_2}$.
This is a rough sketch of the locations where $k$-th type objects are placed.
However, each object could have different attributes such as different style or color.
Hence, it goes through the attribute stage where the tensor is transformed by a SPADE layer \cite{park2019semantic, huh2019feedback} with the masked attribute map $O^k\odot A^k$ given as the contextual information. 
It is further fed through a few transposed convolution layers, and finally goes through an attention process similar to the rough sketch stage where concatenated components goes through a spatial softmax to get fine masks.
The intuition is that after drawing individual objects, it needs to decide the ``depth" ordering of the objects to be drawn in order to account for occlusions.
Let us denote the fine mask as $\eta^k$ and the final tensor as $X^k$.
After this process, the final image is obtained by summing up all components, $x = \sum^K_{k=1} \eta^k \odot X^k$. 
Therefore, the architecture of our neural rendering engine encourages it to extract different information from the memory vector and the hidden state with the help of temporal cycle loss (Section \ref{sec:cycle_loss}).
We also introduce a version with more capacity that can produce higher quality images in Section A.5 of the supplementary materials.

\subsection{Training \Name}
\label{sec:losses}
\vspace{-1mm}

Adversarial training has been successfully employed for image and video synthesis tasks.
\Name leverages adversarial training to learn environment dynamics and to produce realistic temporally coherent simulations.
For certain cases where long-term consistency is required, we propose temporal cycle loss that disentangles static and dynamic components to learn to remember what it has generated.

\vspace{-2mm}
\subsubsection{Adversarial Losses}
\vspace{-1mm}

There are three main components: single image discriminator, action discriminator, and temporal discriminator.

\textbf{Single image discriminator:} \enspace
To ensure each generated frame is realistic, the single image discriminator and \Name simulator play an adversarial game.

\textbf{Action-conditioned discriminator:} \enspace
\Name has to reflect the actions taken by the agent faithfully.
We give three pairs to the action-conditioned discriminator: ($x_t, x_{t+1}, a_t$), ($x_t, x_{t+1}, \bar{a}_t$) and ($\hat{x}_t, \hat{x}_{t+1}, a_t$). 
$x_t$ denotes the real image, $\hat{x}_t$ the generated image, and $\bar{a}_t \in \displaystyle \mathcal{A}$ a sampled negative action $\bar{a}_t \neq a_t$.
The job of the discriminator is to judge if two frames are consistent with respect to the action.
Therefore, to fool the discriminator, \Name has to produce realistic future frame that reflects the action.

\textbf{Temporal discriminator:} \enspace
Different entities in an environment can exhibit different behaviors, and also appear or disappear in partially observed states.
To simulate a temporally consistent environment, one has to take past information into account when generating the next states.
Therefore, we employ a temporal discriminator that is implemented as 3D convolution networks.
It takes several frames as input and decides if they are a real or generated sequence.

Since conditional GAN architectures \cite{mirza2014conditional} are known for learning simplified distributions ignoring the latent code \cite{yang2019diversity, salimans2016improved}, we add information regularization \cite{chen2016infogan} that maximizes the mutual information $I(z_t, \phi(x_{t}, x_{t+1}))$ between the latent code $z_t$ and the pair $(x_t, x_{t+1})$.
To help the action-conditioned discriminator, we add a term that minimizes the cross entropy loss between $a_t$ and $a^{pred}_t = \psi(x_{t+1}, x_t)$.
Both $\phi$ and $\psi$ are MLP that share layers with the action-conditioned discriminator except for the last layer.
Lastly, we found adding a small reconstruction loss in image and feature spaces helps stabilize the training (for feature space, we reduce the distance between the generated and real frame's single image discriminator features).
A detailed descriptions are provided in the supplementary material.

\vspace{-2mm}
\subsubsection{Cycle Loss}
\label{sec:cycle_loss}

RNN based generators are capable of keeping track of the recent past to generate coherent frames.
However, it quickly forgets what happened in the distant past since it is encouraged simply to produce realistic next observation.
To ensure long-term consistency of static elements, we leverage the memory module and the rendering engine to disentangle static elements from dynamic elements.

After running through some time steps $T$, the memory block $\mathcal{M}$ is populated with information from the dynamics engine.
Using the memory location history $\alpha_t$, we can retrieve the memory vector $\hat{m}_t$ which could be different from $m_t$ if the content at the location $\alpha_t$ has been modified.
Now, $c = \{\hat{m}_t, \mathbf{0}\}$ is passed to the rendering engine to produce $X^{\hat{m}_t}$ where $\mathbf{0}$ is the zero vector and $X^{\hat{m}_t}$ is the output component corresponding to $\hat{m}_t$.
We use the following loss: 
\begin{equation}
	L_{cycle} = \sum_t^T||X^{m_t} - X^{\hat{m}_t}||
\end{equation}
As dynamic elements (\eg moving ghosts in Pacman) do not stay the same across time, the engine is encouraged to put static elements in the memory vector to reduce $L_{cycle}$.
Therefore, long-term consistency is achieved.

To prevent the trivial solution where the model tries to ignore the memory component, we use a regularizer that minimizes the sum of all locations in the fine mask $\min\sum\eta^h$ from the hidden state vector so that $X^{m_t}$ has to contain content.
Another trivial solution is if shift kernels for all actions are learned to never be in the opposite direction of each other. 
In this case, $\hat{m}_t$ and $m_t$ would always be the same because the same memory location will never be revisited.
Therefore, we put a constraint that for actions $a$ with a negative counterpart $\hat{a}$ (\eg \textit{Up} and \textit{Down}), $\hat{a}$'s shift kernel $\mathcal{K}(\hat{a})$ is equal to horizontally and vertically flipped $\mathcal{K}(a)$.
Since most simulators that require long-term consistency involves navigation tasks, it is trivial to find such counterparts.

\vspace{-2mm}
\subsubsection{Training Scheme}
\Name is trained end-to-end.
We employ a warm-up phase where real frames are fed into the dynamics engine for the first few epochs,
and slowly reduce the number of real frames to 1 (the initial frame $x_0$ is always given).
We use 18 and 32 frames for training \Name on Pacman and VizDoom environments, respectively.


\begin{figure}
\vspace{-1mm}
\begin{center}
\includegraphics[width=\linewidth]{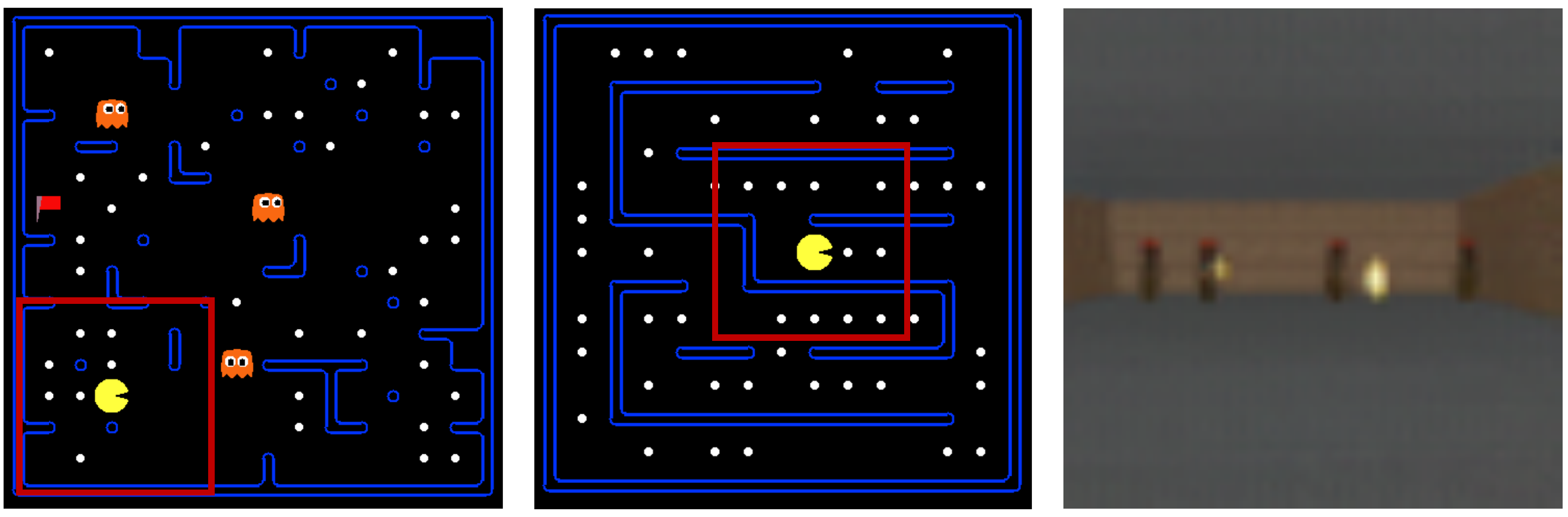}
\end{center}
\vspace{-3mm}
\caption{\footnotesize Samples from datasets studied in this work.
         For Pacman and Pacman-Maze, training data consists of partially observed states, shown in the red box.
          Left: Pacman, Center: Pacman-Maze, Right: VizDoom}
\label{fig:dataset_example}
\vspace{-2mm}
\end{figure}

\begin{figure*}[!hbt]
    \vspace{-2mm}
    \begin{center}
        \includegraphics[width=\textwidth]{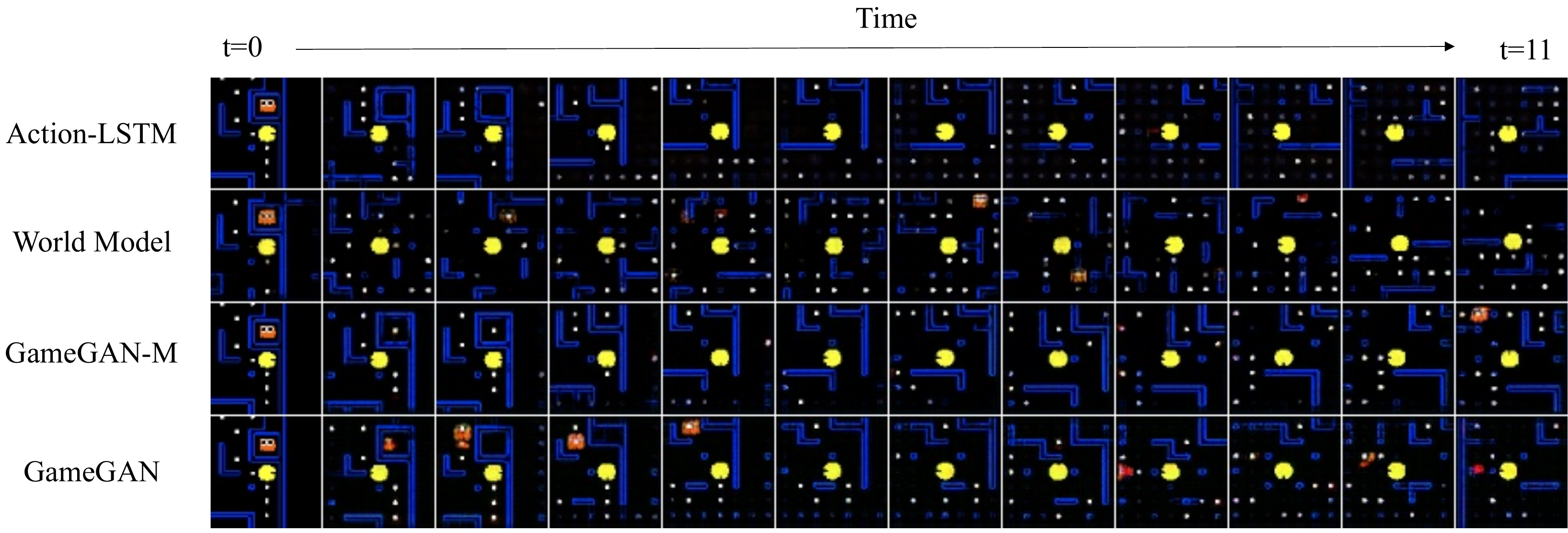}
    \end{center}
    \vspace{-3mm}
    \caption{
        Rollout of models from the same initial screen.
        Action-LSTM trained with reconstruction loss produces frames without refined details (\eg foods).
        World Model has difficulty keeping temporal consistency, resulting in occasional significant discontinuities.
        \Name can produce consistent simulation.
    }
    \label{fig:pacman_rollout}
    \vspace{-2mm}
\end{figure*}

\section{Experiments}
\label{sec:experiments}

We present both qualitative and quantitative experiments.
We mainly consider four models:
1) Action-LSTM: model trained only with reconstruction loss which is in essence similar to~\cite{chiappa2017recurrent},
2) World Model~\cite{ha2018recurrent},
3) GameGAN-M: our model without the memory module and with the simple rendering engine,
and 4) GameGAN: the full model with the memory module and the rendering engine for disentanglement.
Experiments are conducted on the following three datasets (Figure \ref{fig:dataset_example}):

\begin{figure*}[!hbt]
\begin{center}
\includegraphics[width=\textwidth]{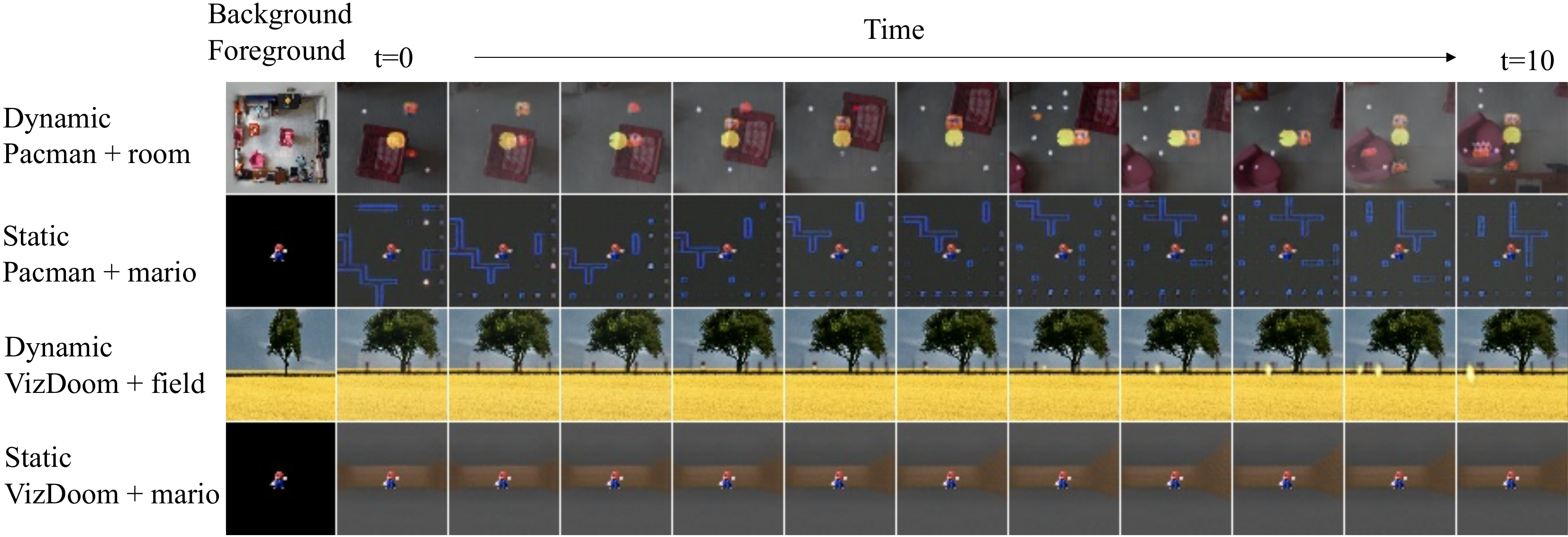}
\end{center}
\vspace{-4mm}
\caption{\Name on Pacman and VizDoom with swapping background/foreground with random images.}
\label{fig:swap_rollout}
\vspace{-5mm}
\end{figure*}

\textbf{Pacman:} \enspace
We use a modified version of the Pacman game\footnote{http://ai.berkeley.edu/project\_overview.html} in which the Pacman agent observes an egocentric 7x7 grid from the full 14x14 environment.
The environment is randomly generated for each episode.
This is an ideal environment to test the quality of a simulator since it has both deterministic (\eg, game rules \& view-point shift) and highly stochastic components (\eg, game layout of foods and walls; game dynamics with moving ghosts).
Images in the episodes are 84x84 and the action space is $\mathcal{A}= \{left, right, up, down, stay\}$.
45K episodes of length greater than or equal to 18 are extracted and 40K are used for training.
Training data is generated by using a trained DQN~\cite{mnih2013playing} agent that observes the full environment with high entropy to allow exploring diverse action sequences.
Each episode consists of a sequence of 7x7 Pacman-centered grids along with actions.

\textbf{Pacman-Maze:} \enspace
This game is similar to Pacman except that it does not have ghosts, and its walls are randomly generated from a maze-generation algorithm, thus are structured better.
The same number of data is used as Pacman.

\textbf{Vizdoom:} \enspace
We follow the experiment set-up of Ha and Schmidhuber \cite{ha2018recurrent} that uses \textit{takecover} mode of the VizDoom platform \cite{kempka2016vizdoom}. 
Training data consists of 10k episodes extracted with random policy.
Images in the episodes are 64x64 and the action space is $\mathcal{A}= \{left, right, stay\}$

\vspace{-1mm}
\subsection{Qualitative Evaluation}
\vspace{-1mm}

Figure~\ref{fig:pacman_rollout} shows rollouts of different models on the Pacman dataset.
Action-LSTM, which is trained only with reconstruction loss, produces blurry images as it fails to capture the multi-modal future distribution, and the errors accumulate quickly.
World model \cite{ha2018recurrent} generates realistic images for VizDoom, but it has trouble simulating the highly stochastic Pacman environment.
In particular, it sometimes suffers from large unexpected discontinuities (\eg $t=0$ to $t=1$).
On the other hand, \Name produces temporally consistent and realistic sharp images.
\Name consists of only a few convolution layers to roughly match the number of parameters of World Model.
We also provide a version of \Name that can produce higher quality images in the supplementary materials Section A.5.

\begin{figure}
    \begin{center}
        \includegraphics[width=8.5cm]{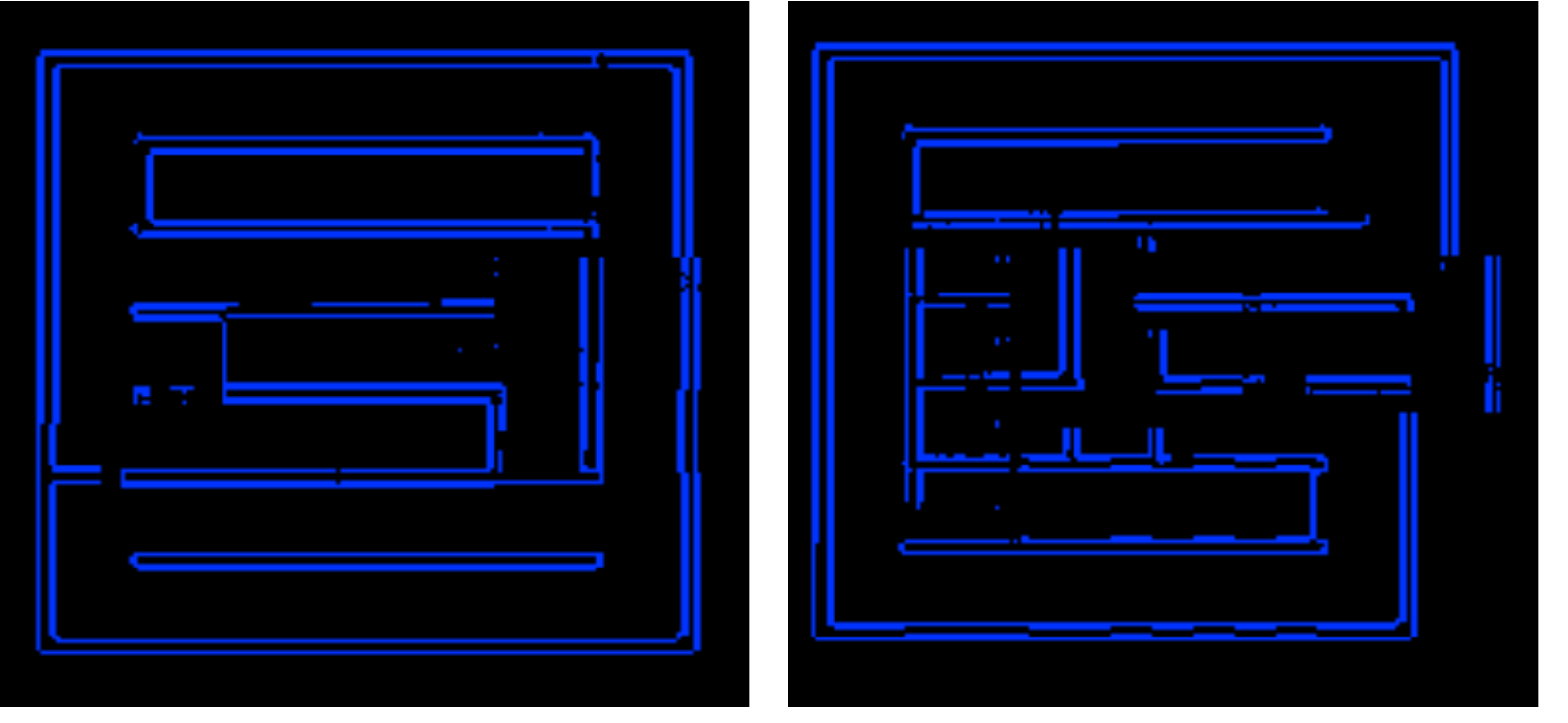}
    \end{center}
    \vspace{-3mm}
    \caption{Generated mazes by traversing with a pacman agent on \Name model.
    Most mazes are realistic.
    Right shows a failure case that does not close the loop correctly.}
    \label{fig:generated_maze}
\vspace{-2mm}
\end{figure}

\textbf{Disentangling static \& dynamic elements:} \enspace
Our \Name with the memory module is trained to disentangle static elements from dynamic elements. 
Figure \ref{fig:disentangle_example} shows how walls from the Pacman environment and the room from the VizDoom environment are separated from dynamic objects such as ghosts and fireballs.
With this, we can make interesting environments in which each element is swapped with other objects.
Instead of the depressing room of VizDoom, enemies can be placed in the user's favorite place, or alternatively have Mario run around the room (Figure~\ref{fig:swap_rollout}).
We can swap the background without having to modify the code of the original games.
Our approach treats games as a black box and learns to reproduce the game, allowing us to easily modify it.
Disentangled models also open up many promising future directions that are not possible with existing models.
One interesting direction would be learning multiple disentangled models and swapping certain components.
As the dynamics engine learns the rules of an environment and the rendering engine learns to render images,
simply learning a linear transformation from the hidden state of one model to make use of the rendering engine of the other could work.

\textbf{Pacman-Maze generation:} \enspace
\Name on the Pacman-Maze produces a partial grid at each time step which can be connected to generate the full maze.
It can generate realistic walls, and as the environment is sufficiently small, \Name also learns the rough size of the map and correctly draws the rectangular boundary in most cases.
One failure case is shown in the bottom right corner of Figure \ref{fig:generated_maze}, that fails to close the loop.

\begin{figure*}[hbt!]
    \vspace{-3mm}
\begin{center}
\includegraphics[width=\textwidth]{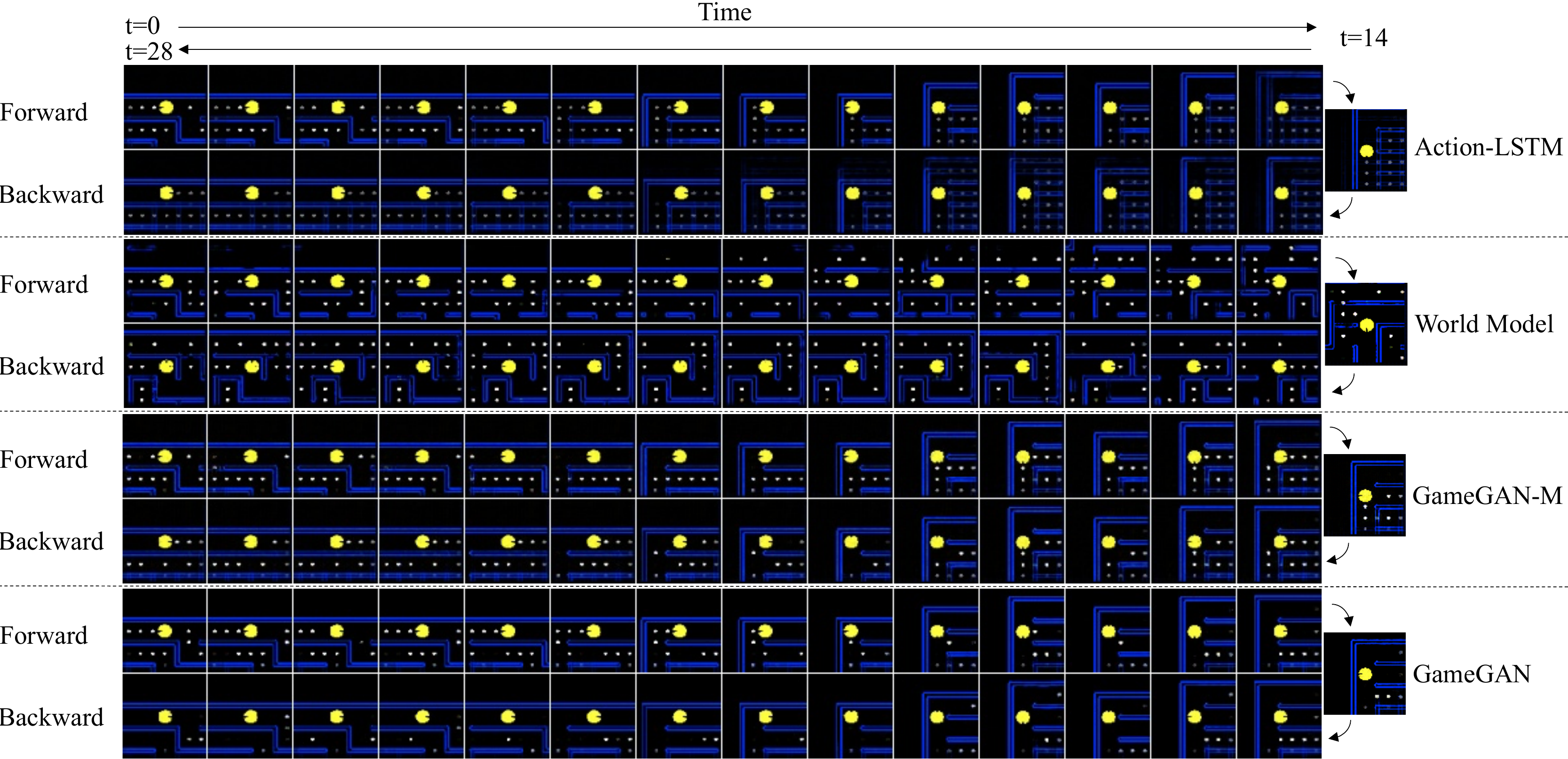}
\end{center}
\vspace{-3mm}
    \caption{Come-back-home task rollouts.
The forward rows show the path going from the initial position to the goal position.
The backward rows show the path coming back to the initial position.
Only the full \Name can successfully recover the initial position. }
\label{fig:pacman_maze_rollout}
\vspace{-3mm}
\end{figure*}

\begin{figure}[h]

\begin{center}
\includegraphics[width=6.5cm]{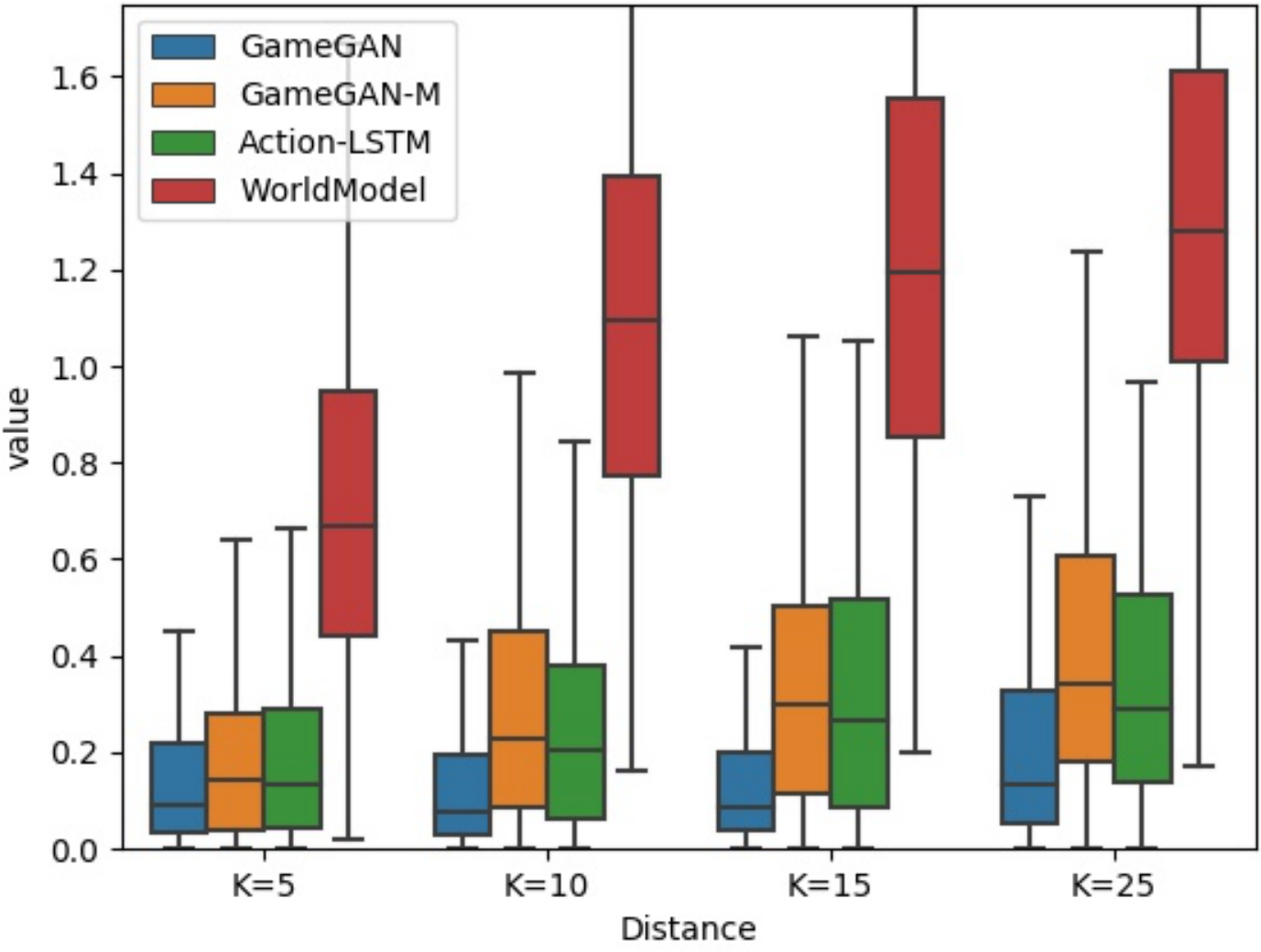}
\end{center}
\vspace{-4mm}

\caption{Box plot for Come-back-home metric. Lower is better. As a reference, a pair of randomly selected frames from the same episode gives a score of 1.17 $\pm$ 0.56 }
\label{fig:cbh_boxplot}
\vspace{-4mm}
\end{figure}

\subsection{Task 1: Training an RL Agent}
\label{sec:exp_rl}
Quantitatively measuring environment quality is challenging as the future is multi-modal, and the ground truth future does not exist.
One way of measuring it is through learning a reinforcement learning agent inside the simulated environment and testing the trained agent in the real environment.
The simulated environment should be sufficiently close to the real one to do well in the real environment.
It has to learn the dynamics, rules, and stochasticity present in the real environment.
The agent from the better simulator that closely resembles the real environment should score higher.
We note that this is closely related to model-based RL.
Since \Name do not internally have a mechanism for denoting the game score, we train an external classifier.
The classifier is given $N$ previous image frames and the current action to produce the output (\eg Win/Lose).

\textbf{Pacman:} \enspace
For this task, the Pacman agent has to achieve a high score by eating foods (+0.5 reward) and capturing the flag (+1 reward).
It is given -1 reward when eaten by a ghost, or the maximum number of steps (40) are used.
Note that this is a challenging partially-observed reinforcement learning task where the agent observes 7x7 grids.
The agents are trained with A3C \cite{mnih2016asynchronous} with an LSTM component.

\textbf{VizDoom:} \enspace
We use the Covariance Matrix Adaptation Evolution Strategy \cite{hansen2001completely} to train RL agents.
Following \cite{ha2018recurrent}, we use the same setting with corresponding simulators.

\vspace{-2mm}
\begin{table}[h!]
\centering\begin{tabular}{ |p{2.6cm}||p{1.9cm}|p{1.9cm}|  }
 \hline
  & Pacman & VizDoom\\
 \hline
 Random Policy & -0.20 $\pm$ 0.78 & 210 $\pm$ 108 \\
 Action-LSTM\cite{chiappa2017recurrent}  & -0.09 $\pm$ 0.87   & 280 $\pm$ 104   \\
 WorldModel\cite{ha2018recurrent}   & 1.24 $\pm$ 1.82   &  1092 $\pm$ 556 \\
 \Name$-M$   & 1.99 $\pm$ 2.23   &   724 $\pm$ 468 \\
 \Name  &   1.13 $\pm$ 1.56 &  765 $\pm$ 482 \\
 \hline
\end{tabular}
\vspace{1mm}
\caption{
Numbers are reported as mean scores $\pm$ standard deviation. Higher is better.
For Pacman, an agent trained in real environment achieves 3.02 $\pm$ 2.64 which can be regarded as the upper bound.
VizDoom is considered solved when a score of 750 is achieved.}
\label{table:result_rl}
\vspace{-2mm}
\end{table}

\begin{figure*}[!htbp]
\begin{center}
\includegraphics[width=\textwidth]{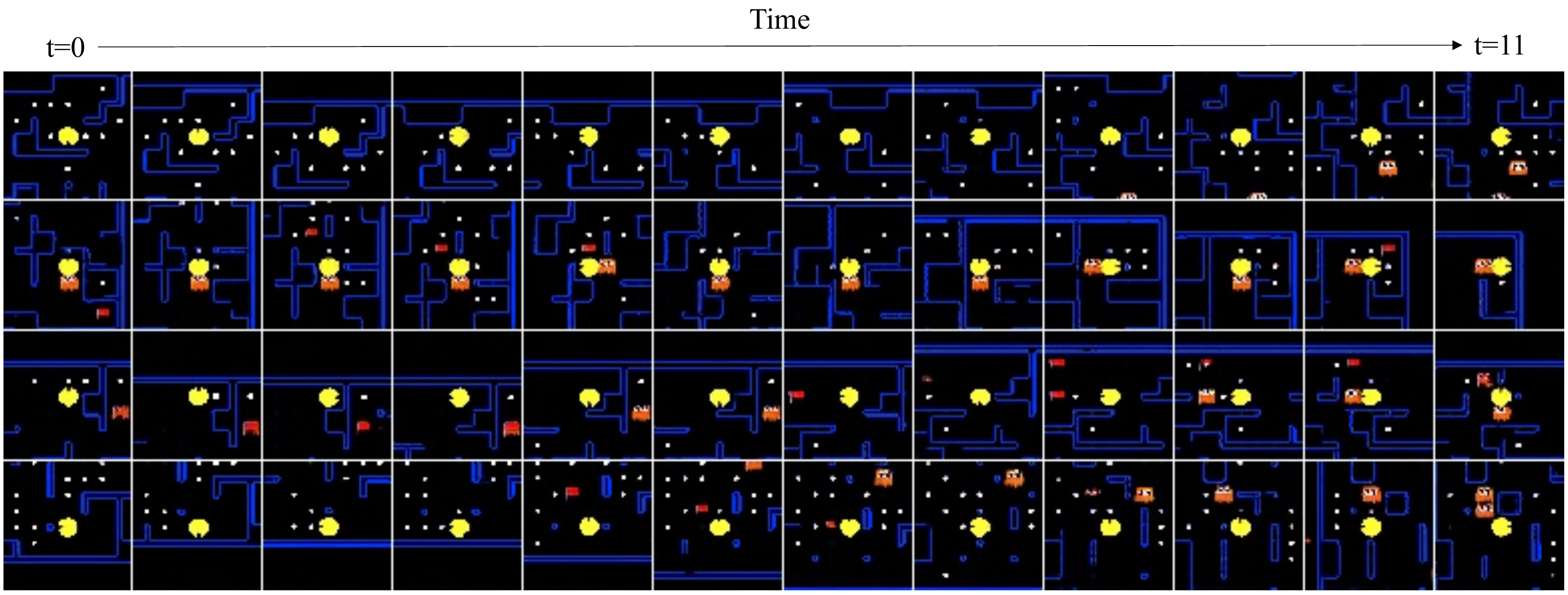}
\end{center}
\caption{Pacman rollouts with a higher-capacity \Name described in Section \ref{sec:supp_more_cap}.
        The image quality is visibly better than the simpler \Name model shown in Figure \ref{fig:pacman_rollout}.}
\label{fig:res_pacman_rollout}
\end{figure*}

Table \ref{table:result_rl} shows the results.
For all experiments, scores are calculated over 100 test environments, and we report the mean scores along with standard deviation.
Agents trained in Action-LSTM simulator performs similar to the agents with random policy, indicating the simulations are far from the real ones.
On Pacman, GameGAN-M shows the best performance while \Name and WorldModel have similar scores.
VizDoom is considered solved when a score of 750 is achieved, and \Name solves the game.
Note that World Model achieves a higher score, but \Name is the first work trained with a GAN framework that solves the game.
Moreover, \Name can be trained end-to-end, unlike World Model that employs sequential training for stability.
One interesting observation is that \Name shows lower performance than GameGAN-M on the Pacman environment.
This is due to having additional complexity in training the model where the environments do not need long-term consistency for higher scores.
We found that optimizing the GAN objective while training the memory module was harder, and this attributes to RL agents exploiting the imperfections of the environments to find a way to cheat.
In this case, we found that \Name sometimes failed to prevent agents from walking through the walls while GameGAN-M was nearly perfect.
This led to RL agents discovering a policy that liked to hit the walls, and in the real environment, this often leads to premature death.
In the next section, we show how having long-term consistency can help in certain scenarios.

\vspace{-1mm}
\subsection{Task 2: Come-back-home}
\label{sec:exp_comebackhome}
\vspace{-1mm}

This task evaluates the long-term consistency of simulators in the Pacman-Maze environment.
The Pacman starts at a random initial position $(x_A, y_A)$ with state $s$.
It is given $K$ random actions $(a_1,...,a_K)$, ending up in position $(x_B, y_B)$.
Using the reverse actions $(\hat{a}_K,...,\hat{a}_1)$(\eg $a_k = Down, \hat{a}_k = Up$) , it comes back to the initial position $(x_A, y_A)$, resulting in state $\hat{s}$.
Now, we can measure the distance $d$ between $\hat{s}$ and $s$ to evaluate long-term consistency ($d$ = 0 for the real environment).
As elements other than the wall (\eg food) could change, we only compare the walls of $\hat{s}$ and $s$.
Hence, $s$ is an 84x84 binary image whose pixel is 1 if the pixel is blue.
We define the metric $d$ as

\vspace{-1mm}
\begin{equation}
d = \frac{\mathrm{sum}(\mathrm{abs}(s - \hat{s}))}{\mathrm{sum}(s) + 1}
\end{equation}
where $\mathrm{sum}()$ counts the number of 1s.
Therefore, $d$ measures the ratio of the number of pixels changed to the initial number of pixels.
Figure \ref{fig:cbh_boxplot} shows the results.
We again observe occasional large discontinuities in World Model that hurts the performance a lot.
When $K$ is small, the differences in performance are relatively small.
This is because other models also have short-term consistency realized through RNNs.
However, as $K$ becomes larger, \Name with memory module steadily outperforms other models, and the gaps become larger, indicating \Name can make efficient use of the memory module.
Figure \ref{fig:pacman_maze_rollout} shows the rollouts of different models in the Pacman-Maze environment.
As it can be seen, models without the memory module do not remember what it has generated before.
This shows \Name opens up promising directions for not only game simulators, but as a general environment simulator that could mimic the real world.

\vspace{-1mm}
\section{Conclusion}
\vspace{-1mm}

We propose \Name which  leverages adversarial training to learn to simulate games.
\Name is trained by observing screenplay along with user's actions and does not require access to the game's logic or engine.
\Name features a new memory module to ensure long-term consistency and is trained to separate static and dynamic elements.
Thorough ablation studies showcase the modeling power of GameGAN.
In future works, we aim to extend our model to capture more complex real-world environments.

\section*{Acknowledgments}
We thank Bandai-Namco Entertainment Inc. for providing the official version of Pac-Man for training.
We also thank Ming-Yu Liu and Shao-Hua Sun for helpful discussions.

{\small
\bibliographystyle{ieee_fullname}
\bibliography{egbib}
}
\clearpage







\newpage

\begin{large}
\textbf{Supplementary Material}
\end{large}
\\
\appendix

\addcontentsline{toc}{section}{Appendices}
\renewcommand{\thesection}{\Alph{section}}

We provide detailed descriptions of the model architecture (Section A), training scheme (Section B), and additional figures (Section C).

\section{Model Architecture}
\label{sec:supp_model}
We provide architecture details of each module described in Section 3.
We adopt the following notation for describing modules:

\textbf{Conv2D(a,b,c,d):}  2D-Convolution layer with output channel size \textbf{a}, kernel size \textbf{b}, stride \textbf{c}, and padding size \textbf{d}.

\textbf{Conv3D(a,b,c,d,e,f,g):}  3D-Convolution layer with output channel size \textbf{a}, temporal kernel size \textbf{b}, spatial kernel size \textbf{c}, temporal stride \textbf{d},
spatial stride \textbf{e}, temporal padding size \textbf{f}, and spatial padding size \textbf{g}.

\textbf{T.Conv2D(a,b,c,d,e):}  Transposed 2D-Convolution layer with output channel size \textbf{a}, kernel size \textbf{b}, stride \textbf{c}, padding size \textbf{d}, and output padding size \textbf{e}.

\textbf{Linear(a):}  Linear layer with output size \textbf{a}.

\textbf{Reshape(a):}  Reshapes the input tensor to output size \textbf{a}.

\textbf{LeakyReLU(a):}  Leaky ReLU function with slope \textbf{a}.

\begin{figure}[bth!]

\begin{center}
\includegraphics[width=0.8\linewidth]{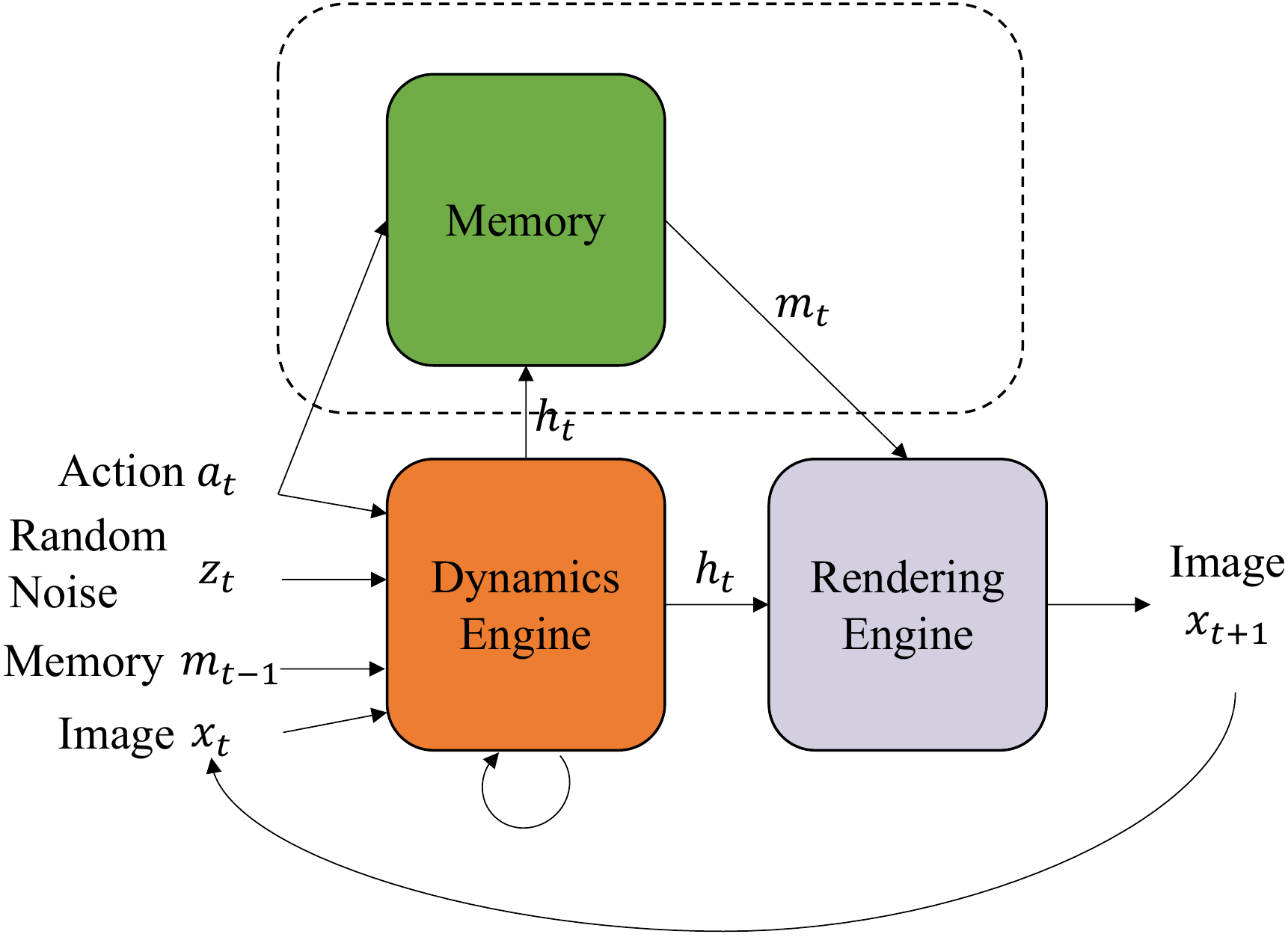}
\end{center}
\vspace{-4mm}
\caption{\footnotesize Overview of GameGAN: The \emph{dynamics engine} takes $a_t, z_t,$ and $x_t$ as input to update the hidden state at time $t$.
        Optionally, it can write to and read from the external \emph{memory module} (in the dashed box).
        Finally, the \emph{rendering engine} is used to decode the output image $x_{t+1}$.
        The whole system is trained end-to-end. All modules are neural networks.}
\label{fig:supp_overview}
\end{figure}

\subsection{Dynamics Engine}
\label{sec:supp_dynamics_engine}
The input action $a \sim \displaystyle \mathcal{A}$ is a one-hot encoded vector.
For Pacman environment, we define $\mathcal{A}= \{left, right, up, down, stay\}$, and for the counterpart actions $\hat{a}$ (see Section 3.4.2),
we define $\hat{left} = right$, $\hat{up} = down$, and vise versa.
The images $x$ have size $84\times84\times3$.
For VizDoom, $\mathcal{A}= \{left, right, stay\}$, and  $\hat{left} = right$, $\hat{right} = left$.
The images $x$ have size $64\times64\times3$.

At each time step $t$, a 32-dimensional stochastic variable $z_t$ is sampled from the standard normal distribution $\displaystyle \mathcal{N} (0; I)$.
Given the history of images $x_{1:t}$ along with $a_t$ and $z_t$, \Name predicts the next image $x_{t+1}$.

For completeness, we restate the computation sequence of the dynamics engine (Section 3.1) here.

\begin{equation}
v_t = h_{t-1} \odot \mathcal{H}(a_t, z_t, m_{t-1})
\end{equation}
\begin{equation}
 s_t = \mathcal{C}(x_t)
\end{equation}
\begin{equation}
    \begin{gathered}
        i_t = \sigma(W^{iv} v_t + W^{is} s_t),
        f_t = \sigma(W^{fv} v_t + W^{fs} s_t), \\
        o_t = \sigma(W^{ov} v_t + W^{os} s_t)
    \end{gathered}
\end{equation}
\begin{equation}
c_t = f_t \odot c_{t-1} + i_t \odot \tanh(W^{cv}v_t + W^{cs} s_t)
\end{equation}
\begin{equation}
h_t = o_t \odot \tanh(c_t)
\end{equation}
where $h_t, a_t, z_t, c_t, x_t$ are the hidden state, action, stochastic variable, cell state, image at time step $t$.
$\odot$ denotes the hadamard product.

The hidden state of the dynamics engine is a 512-dimensional vector. Therefore, $h_t, o_t, c_t, f_t, i_t, o_t \in  \mathbb{R}^{512}$.

$\mathcal{H}$ first computes embeddings for each input. Then it concatenates and passes them through two-layered MLP: [Linear(512),LeakyReLU(0.2),Linear(512)].
$h_{t-1}$ can also go through a linear layer before the hadamard product in eq.13, if the size of hidden units differ from 512.

$\mathcal{C}$ is implemented as a 5 (for 64$\times$64 images) or 6 (for 84$\times$84 images) layered convolutional networks, followed by a linear layer:

\begin{center}
\vspace{-3mm}
\begin{tabular}{ |c | c|}
 \hline
  Pacman & VizDoom\\
 \hline\hline

  Conv2D(64, 3, 2, 0) & Conv2D(64, 4, 1, 1) \\
 \hline
 LeakyReLU(0.2)  & LeakyReLU(0.2) \\
 \hline

  Conv2D(64, 3, 1, 0)     & Conv2D(64, 3, 2, 0) \\
 \hline
 LeakyReLU(0.2) & LeakyReLU(0.2) \\
 \hline

  Conv2D(64, 3, 2, 0)    & Conv2D(64,3, 2, 0) \\
 \hline
 LeakyReLU(0.2) & LeakyReLU(0.2) \\
 \hline

  Conv2D(64, 3, 1, 0)    & Conv2D(64, 3, 2, 0) \\
 \hline
 LeakyReLU(0.2) & LeakyReLU(0.2) \\
 \hline
 Conv2D(64, 3, 2, 0) & Reshape(7*7*64) \\
 \hline
 LeakyReLU(0.2) & Linear(512) \\
 \hline
 Reshape(8*8*64) &  \\
 \hline
 Linear(512) &  \\

 \hline
\end{tabular}

\end{center}

\subsection{Memory Module}
\label{sec:supp_memory_module}
\begin{figure}[hbt!]
\begin{center}
\includegraphics[width=8.5cm]{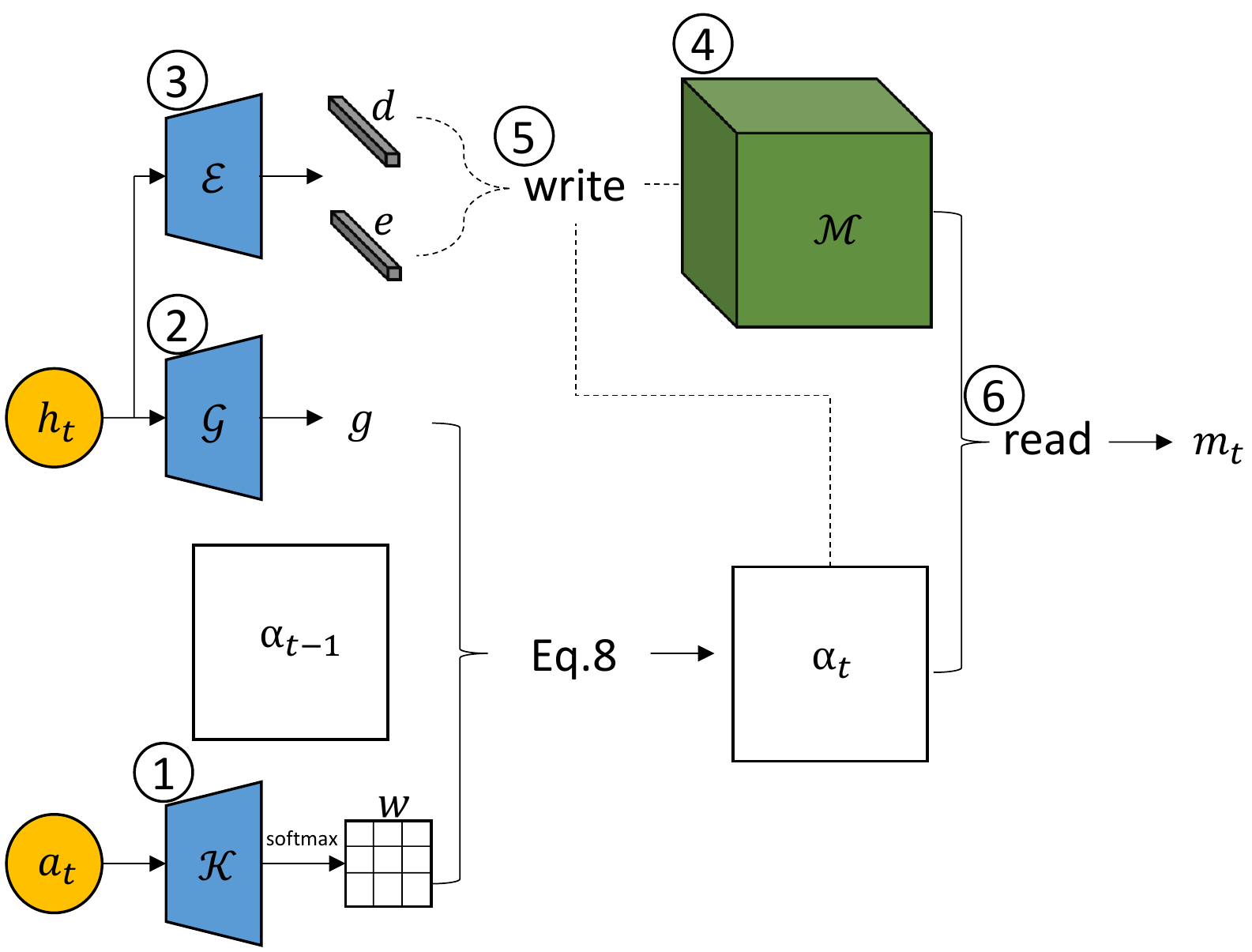}
\end{center}
\caption{\footnotesize \emph{Memory Module}. Descriptions for each numbered circle are provided in the text.}
\label{fig:supp_memory}
\end{figure}

For completeness, we restate the computation sequence of the memory module (Section 3.2) here.

\begin{equation}
w = \mathrm{softmax}(\mathcal{K}(a_t)) \in \mathbb{R}^{3\times 3}
\end{equation}
\begin{equation}
g = \mathcal{G}(h_t) \in \mathbb{R}
\end{equation}
\begin{equation}
\alpha_{t} = g \cdot \mathrm{Conv2D}(\alpha_{t-1}, w) + (1-g)\cdot \alpha_{t-1}
\end{equation}
\begin{equation}
\mathcal{M} = \mathrm{write}(\alpha_t, \mathcal{E}(h_t), \mathcal{M})
\end{equation}
\begin{equation}
m_t = \mathrm{read}(\alpha_t, \mathcal{M})
\end{equation}

See Figure \ref{fig:supp_memory} for each numbered circle.

{\Large \textcircled{\normalsize 1}} $\mathcal{K}$ is a two-layered MLP that outputs a 9 dimensional vector which is softmaxed and reshaped to 3$\times$ 3 kernel $w$:
[Linear(512), LeakyReLU(0.2), Linear(9), Softmax(), Reshape(3,3)].

{\Large \textcircled{\normalsize 2}} $\mathcal{G}$ is a two-layered MLP that outputs a scalar value followed by the sigmoid activation function such that $g \in [0,1]$: [Linear(512),LeakyReLU(0.2),Linear(1),Sigmoid()].

{\Large \textcircled{\normalsize 3}} $\mathcal{E}$ is a one-layered MLP that produces an erase vector $e \in \mathbb{R}^{512}$ and an add vector $d \in \mathbb{R}^{512}$: [Linear(1024), split(512)],
where Split(512) splits the 1024 dimensional vector into two 512 dimensional vectors.
$e$ additionally goes through the sigmoid activation function.

{\Large \textcircled{\normalsize 4}} Each spatial location in the memory block $\mathcal{M}$ is initialized with 512 dimensional random noise $\sim N(0,I)$.
For computational efficiency, we use the block width and height $N=39$ for training and $N=99$ for downstream tasks.
Therefore, we use $39\times39\times512$ blocks for training, and $99\times99\times512$ blocks for experiments.
Note that the shift-based memory module architecture allows any arbitrarily sized blocks to be used at test time.

{\Large \textcircled{\normalsize 5}} $\mathrm{write}$ operation is implemented similar to the Neural Turing Machine \cite{graves2014neural}.
For each location $\mathcal{M}^i$, $\mathrm{write}$ computes:
\begin{equation}
\mathcal{M}^i = \mathcal{M}^i (1-\alpha_t^i \cdot e) + \alpha_t^i \cdot d
\end{equation}
where $i$ denotes the spatial $x,y$ coordinates of the block $\mathcal{M}$.
$e$ is a sigmoided vector which erases information from $\mathcal{M}^i$ when $e = \mathbf{1}$, and $d$ writes new information to $\mathcal{M}^i$.
Note that if the scalar $\alpha_t^i$ is 0 (\ie the location is not being attended), the memory content $\mathcal{M}^i$ does not change.

{\Large \textcircled{\normalsize 6}} $\mathrm{read}$ operation is defined as:
\begin{equation}
m_t = \sum_{i=0}^{N\times N} \alpha_t^i \cdot \mathcal{M}^i
\end{equation}
where $\mathcal{M}^i$ denotes the memory content at location $i$.

\subsection{Rendering Engine}
\label{sec:supp_rendering_engine}

For the simple rendering engine of \Name-M, we first pass the hidden state $h_t$ to a linear layer to make
it a $7\times7\times512$ tensor, and pass it through 5 transposed convolution layers to produce the 3-channel output image $x_{t+1}$:
\begin{tabular}{ |c | c|}
 \hline
  Pacman & VizDoom\\
 \hline\hline

  Linear(512*7*7) & Linear(512*7*7) \\
 \hline
 LeakyReLU(0.2)  & LeakyReLU(0.2) \\
 \hline

  Reshape(7, 7, 512)     & Reshape(7, 7, 512) \\
 \hline
  T.Conv2D(512, 3, 1, 0, 0) & T.Conv2D(512, 4, 1, 0, 0) \\
 \hline
 LeakyReLU(0.2) & LeakyReLU(0.2) \\
 \hline

  T.Conv2D(256, 3, 2, 0, 1) &T.Conv2D(256, 4, 1, 0, 0) \\
 \hline
 LeakyReLU(0.2) & LeakyReLU(0.2) \\
 \hline
 T.Conv2D(128, 4, 2, 0, 0) & T.Conv2D(128, 5, 2, 0, 0) \\
 \hline
 LeakyReLU(0.2) & LeakyReLU(0.2) \\
 \hline
 T.Conv2D(64, 4, 2, 0, 0) & T.Conv2D(64, 5, 2, 0, 0) \\
 \hline
 LeakyReLU(0.2) & LeakyReLU(0.2) \\
 \hline
 T.Conv2D(3, 3, 1, 0, 0) & T.Conv2D(3, 4, 1, 0, 0)  \\

 \hline
\end{tabular}
\vspace{3mm}

\begin{figure}[hbt!]
\begin{center}
\includegraphics[width=8.5cm]{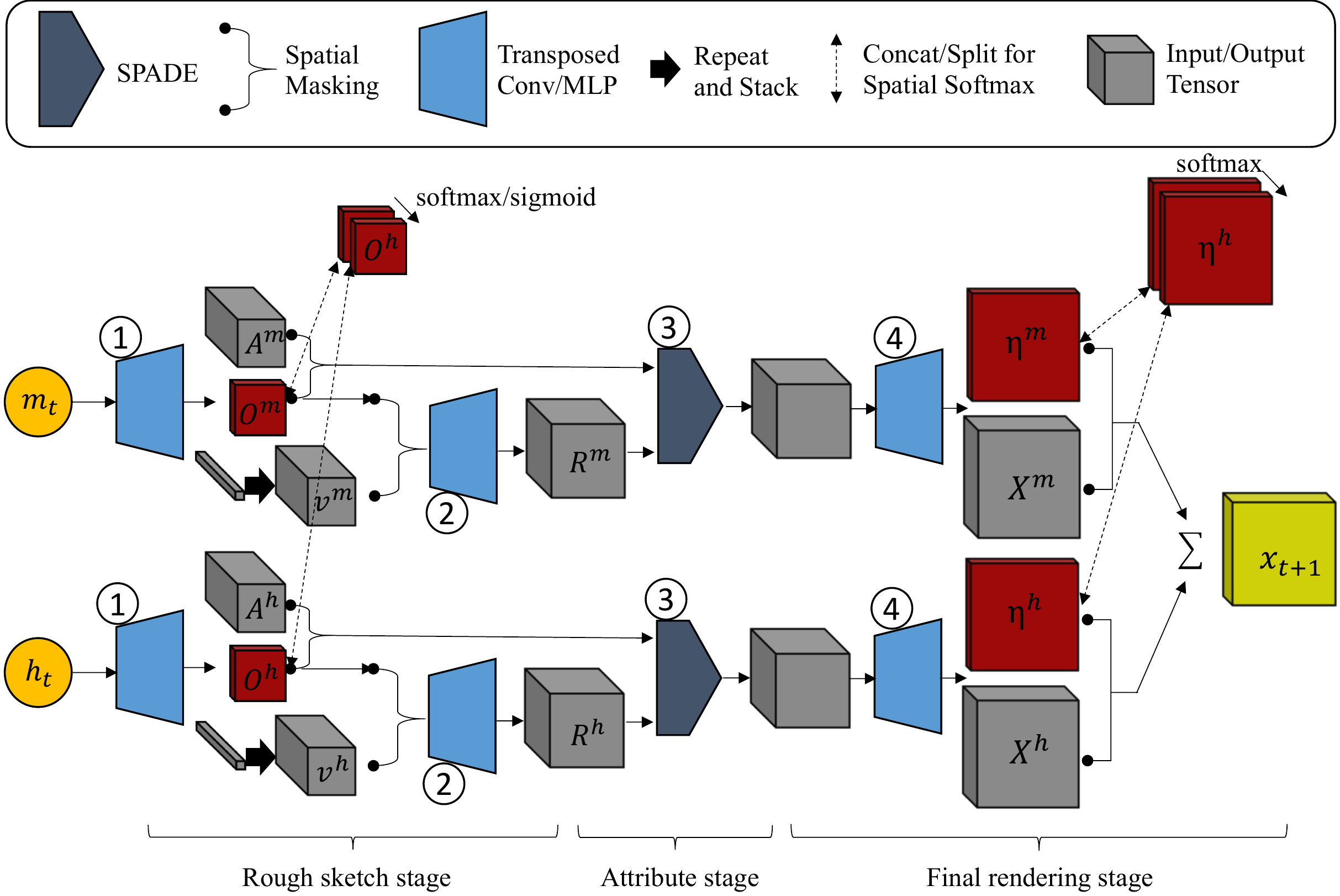}
\end{center}
\caption{\footnotesize \emph{Rendering engine} for disentangling static and dynamic components.  Descriptions for each numbered circle are provided in the text.}
\label{fig:supp_renderer}
\end{figure}

For the specialized disentangling rendering engine that takes a list of vectors $\mathbf{c} = \{c^1,...,c^K\} = \{m_h, h_t\}$ as input, the followings are implemented.
See Figure \ref{fig:supp_renderer} for each numbered circle.

{\Large \textcircled{\normalsize 1}} First, $A^k \in \mathbb{R}^{H_1\times H_1\times D_1}$ and $O^k \in  \mathbb{R}^{H_1\times H_1\times 1}$ are obtained by passing $c^k$ to a linear layer to make it a $3\times3\times128$ tensor,
and the tensor is passed through two transposed convolution layers with filter sizes 3 to produce  $\mathbb{R}^{7\times 7\times 32+1}$ tensor (hence, $H_1 =7, D_1=32$):
\begin{center}
\begin{tabular}{ |c|  }
 \hline
  Pacman \& VizDoom\\
 \hline\hline

 Linear(3*3*128) \\
 \hline
 Reshape(3,3,128) \\
 \hline
 LeakyReLU(0.2) \\
 \hline
 T.Conv2D(512, 3, 1, 0, 0)  \\
 \hline
 LeakyReLU(0.2) \\
 \hline
 T.Conv2D(32+1, 3, 1, 0, 0)  \\
 \hline

\end{tabular}
\end{center}

We split the resulting tensor channel-wise to get $A^k$ and $O^k$.
$v^k$ is obtained from running $c^k$ through one-layered MLP [Linear(32),LeakyReLU(0.2)] and stacking it across the spatial dimension to match the size of $A^k$.

{\Large \textcircled{\normalsize 2}} The rough sketch tensor $R^k$ is obtained by passing $v^k$ masked by $O^k$ (which is either spatially softmaxed or sigmoided) through two transposed convolution layers:
\begin{table}[tbh!]
\centering
\vspace{-3mm}
\begin{tabular}{ |c|c|  }
 \hline
  Pacman & VizDoom\\
 \hline\hline

 T.Conv2D(256, 3, 1, 0, 0) & T.Conv2D(256, 3, 1, 0, 0) \\
 \hline
  LeakyReLU(0.2) & LeakyReLU(0.2) \\
 \hline
 T.Conv2D(128, 3, 2, 0, 1) & T.Conv2D(128, 3, 2, 1, 0) \\
 \hline

\end{tabular}
\vspace{-3mm}
\end{table}

{\Large \textcircled{\normalsize 3}} We follow the architecture of SPADE \cite{park2019semantic} with instance normalization as the normalization scheme.
The attribute map $A^k$ masked by $O^k$ is used as the semantic map that produces the parameters of the SPADE layer, $\gamma$ and $\beta$.

{\Large \textcircled{\normalsize 4}} The normalized tensor goes through two transposed convolution layers:
\begin{center}
\centering
\vspace{0mm}
\begin{tabular}{ |c|c|  }
 \hline
  Pacman & VizDoom\\
 \hline\hline

 T.Conv2D(64, 4, 2, 0, 0) & T.Conv2D(64, 3, 2, 1, 0) \\
 \hline
  LeakyReLU(0.2) & LeakyReLU(0.2) \\
 \hline
 T.Conv2D(32, 4, 2, 0, 0) & T.Conv2D(32, 3, 2, 1, 0) \\
 \hline
  LeakyReLU(0.2) & LeakyReLU(0.2) \\
 \hline

\end{tabular}
\end{center}

From the output tensor of the above, $\eta^k$ is obtained with a single convolution layer [Conv2D(1, 3, 1)] and then is concatenated with other components for spatial softmax.
We also experimented with concatenating $\eta$ and passing them through couple of 1$\times$1 convolution layers before the softmax, and did not observe much difference.
Similarly, $X^k$ is obtained with a single convolution layer [Conv2D(3, 3, 1)].

\subsection{Discriminators}
\label{sec:supp_discriminator}

There are several discriminators used in \Name.
For computational efficiency, we first get an encoding of each frame with a shared encoder:
\begin{center}
\begin{tabular}{ |c|c|  }
 \hline
  Pacman & VizDoom\\
 \hline\hline

 Conv2D(16, 5, 2, 0) & Conv2D(64, 4, 2, 0) \\
 \hline
  BatchNorm2D(16) & BatchNorm2D(64) \\
 \hline
  LeakyReLU(0.2) & LeakyReLU(0.2) \\
 \hline
 Conv2D(32, 5, 2, 0) & Conv2D(128, 3, 2, 0) \\
 \hline
  BatchNorm2D(32) & BatchNorm2D(128) \\
 \hline
  LeakyReLU(0.2) & LeakyReLU(0.2) \\
 \hline
 Conv2D(64, 3, 2, 0) & Conv2D(256, 3, 2, 0) \\
 \hline
  BatchNorm2D(64) & BatchNorm2D(256) \\
 \hline
  LeakyReLU(0.2) & LeakyReLU(0.2) \\
 \hline
 Conv2D(64, 3, 2, 0) & Conv2D(256, 3, 2, 0) \\
 \hline
  BatchNorm2D(64) & BatchNorm2D(256) \\
 \hline
  LeakyReLU(0.2) & LeakyReLU(0.2) \\
 \hline
Reshape(3, 3, 64) & Reshape(3, 3, 256) \\
 \hline

\end{tabular}
\end{center}

\textbf{Single Frame Discriminator} \enspace
The job of the single frame discriminator is to judge if the given single frame is realistic or not.
We use two simple networks for the patch-based ($D_{patch}$) and the full frame ($D_{full}$) discriminators:

\begin{tabular}{ |c|c|  }
 \hline
  $D_{patch}$ & $D_{full}$\\
 \hline\hline

 Conv2D($dim$, 2, 1, 1) & Conv2D($dim$, 2, 1, 0) \\
 \hline
  BatchNorm2D($dim$) & BatchNorm2D($dim$) \\
 \hline
  LeakyReLU(0.2) & LeakyReLU(0.2) \\
 \hline
 Conv2D(1, 1, 2, 1) & Conv2D(1, 1, 2, 1) \\
 \hline

\end{tabular}
\vspace{3mm}

where $dim$ is 64 for Pacman and 256 for VizDoom.
$D_{patch}$ gives 3$\times$3 logits, and $D_{full}$ gives a single logit.

\textbf{Action-conditioned Discriminator} \enspace
We give three pairs to the action-conditioned discriminator $D_{action}$: ($x_t, x_{t+1}, a_t$), ($x_t, x_{t+1}, \bar{a}_t$) and ($\hat{x}_t, \hat{x}_{t+1}, a_t$).
$x_t$ denotes the real image, $\hat{x}_t$ the generated image, and $\bar{a}_t \in \displaystyle \mathcal{A}$ a negative action which is sampled such that $\bar{a}_t \neq a_t$.
The job of the discriminator is to judge if two frames are consistent with respect to the given action.
First, we get an embedding vector for the one-hot encoded action through an embeddinig layer [Linear($dim$)], where $dim=32$ for Pacman and $dim=128$ for VizDoom.
Then, two frames $x_t, x_{t+1}$ are concatenated channel-wise and merged with a convolution layer [Conv2D($dim$, 3, 1, 0), BatchNorm2D($dim$), LeakyReLU(0.2), Reshape($dim$)].
Finally, the action embedding and merged frame features are concatenated together and fed into [Linear($dim$), BatchNorm1D($dim$), LeakyReLU(0.2), Linear(1)], resulting in a single logit.

\textbf{Temporal Discriminator} \enspace
The temporal discriminator $D_{temporal}$ takes several frames as input and decides if they are a real or generated sequence.
We implement a hierarchical temporal discriminator that outputs logits at several levels.
The first level concatenates all frames in temporal dimension and does:

\begin{center}
\begin{tabular}{ |c|  }
 \hline
 Conv3D(64, 2, 2, 1, 1, 0, 0) \\
 \hline
  BatchNorm3D(64) \\
 \hline
  LeakyReLU(0.2) \\
 \hline
 Conv3D(128, 2, 2, 1, 1, 0, 0) \\
 \hline
  BatchNorm3D(128) \\
 \hline
  LeakyReLU(0.2) \\
 \hline

\end{tabular}
\end{center}

The output tensor from the above are fed into two branches.
The first one is a single layer [Conv3D(1, 2, 1, 2, 1, 0, 0)] that produces a single logit.
Hence, this logit effectively looks at 6 frames and judges if they are real or not.
The second branch uses the tensor as an input to the next level:

\begin{table}[tbh!]
\centering
\vspace{-3mm}
\begin{tabular}{ |c|  }
 \hline
 Conv3D(256, 3, 1, 2, 1, 0, 0) \\
 \hline
  BatchNorm3D(256) \\
 \hline
  LeakyReLU(0.2) \\
 \hline

\end{tabular}
\vspace{-3mm}
\end{table}

Now, similarly, the output tensor is fed into two branches.
The first one is a single layer [Conv3D(1, 2, 1, 1, 1, 0, 0)] producing a single logit that effectively looks at 18 frames.
The second branch uses the tensor as an input to the next level:

\begin{table}[tbh!]
\centering
\vspace{-3mm}
\begin{tabular}{ |c|  }
 \hline
 Conv3D(512, 3, 1, 2, 1, 0, 0) \\
 \hline
  BatchNorm3D(512) \\
 \hline
  LeakyReLU(0.2) \\
  \hline
 Conv3D(1, 3, 1, 1, 1, 0, 0) \\
 \hline
\end{tabular}
\vspace{-3mm}
\end{table}
which gives a single logit that has an temporal receptive field size of 32 frames.

The Pacman environment uses two levels (up to 18 frames) and VizDoom uses all three levels.

\begin{figure}
\begin{subfigure}{0.21\textwidth}
\includegraphics[width=\linewidth,height=4.7cm]{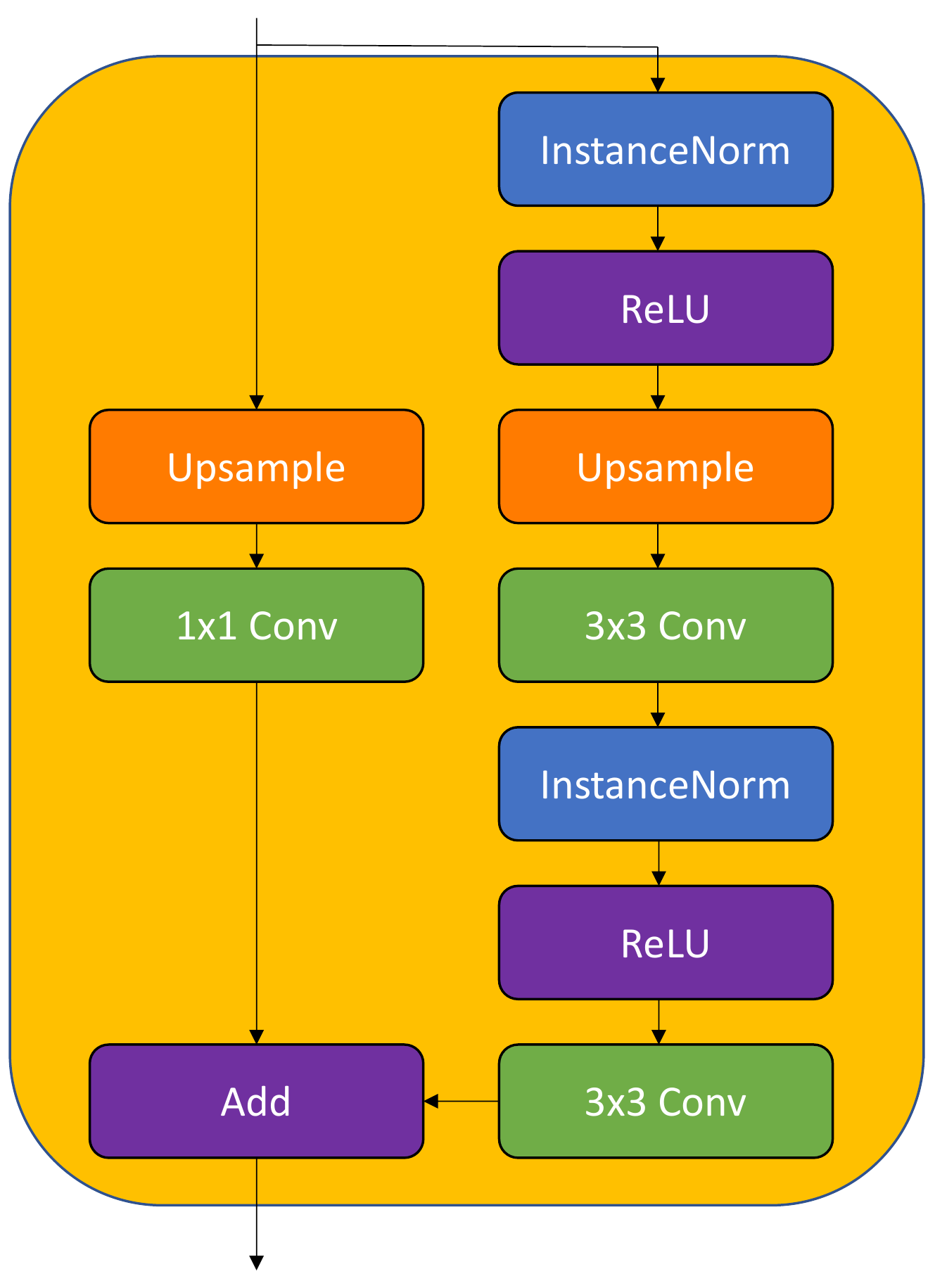}\caption{}
 \label{fig:resblock_g}
\end{subfigure}
\hspace*{\fill} 
\begin{subfigure}{0.21\textwidth}
\includegraphics[width=\linewidth,height=4.7cm]{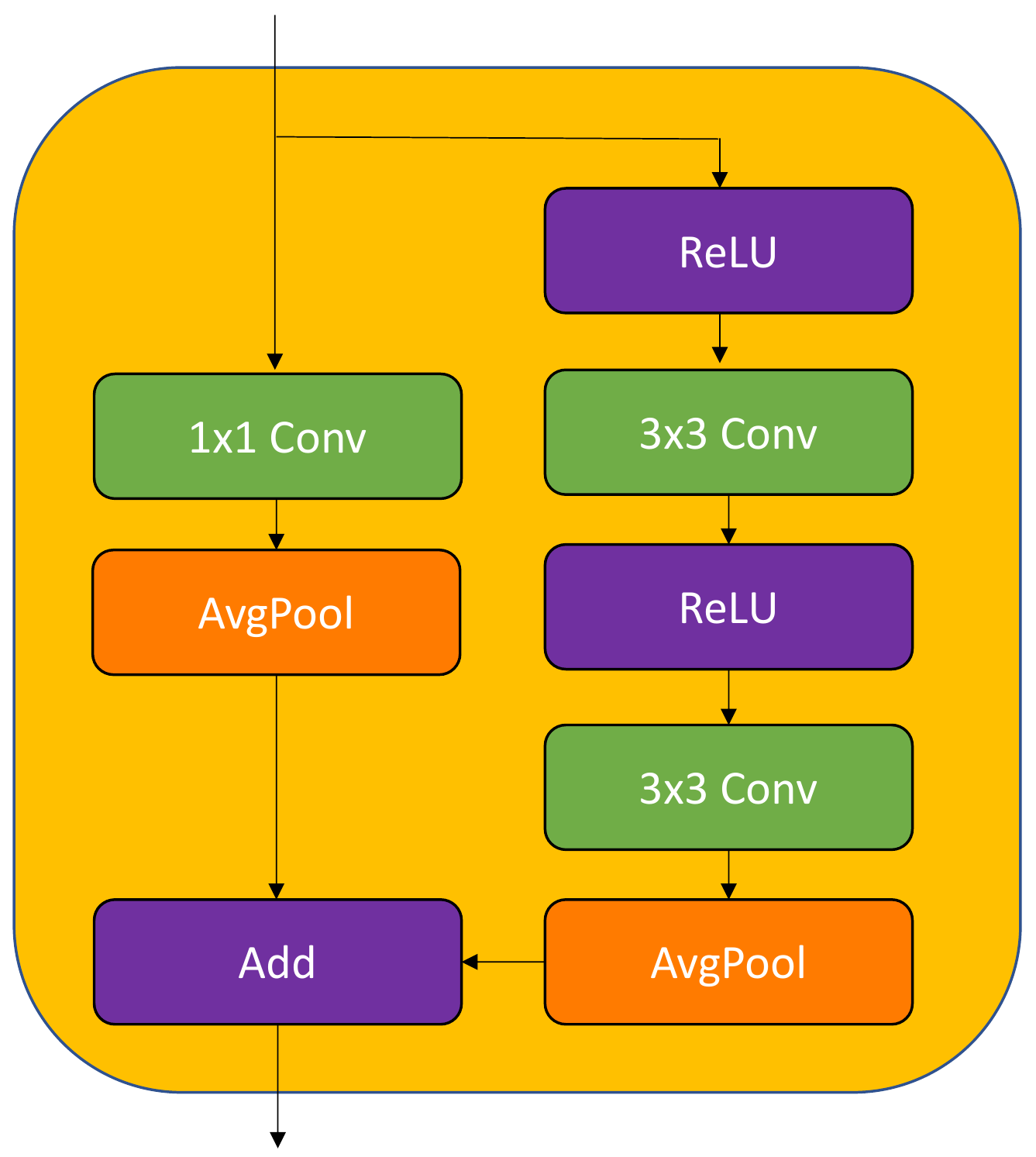}\caption{}
 \label{fig:resblock_d}
\end{subfigure}
\caption{Residual blocks for the higher capacity model, modified and redrawn from Figure 15 of \cite{brock2018large}. (a) A block for rendering engine, (b) A block for discriminator.}
\label{fig:higher_capacity_resblock}
\end{figure}

\subsection{Adding More Capacity}
\label{sec:supp_more_cap}

Both the rendering engine and discriminator described in \ref{sec:supp_rendering_engine} and \ref{sec:supp_discriminator} consist of only a few linear and convolutional layers which can limit \Name's expressiveness.
Motivated by recent advancements in image generation GANs  \cite{brock2018large,park2019semantic}, we experiment with having higher capacity residual blocks.
Here, we highlight key differences compared to the previous sections.
The code will be released for reproducibility.

\textbf{Rendering Engine:}\enspace
The convolutional layers in rendering engine are replaced by residual blocks described in Figure \ref{fig:resblock_g}.
The residual blocks follow the design of \cite{brock2018large} with the difference being that batch normalization layers \cite{ioffe2015batch} are replaced with instance normalization layers \cite{ulyanov2016instance}.
For the specialized rendering engine, having only two object types as in \ref{sec:supp_rendering_engine} could also be a limiting factor.
We can easily add more components by adding to the list of vectors (for example, let $ \mathbf{c} = \{ m_h, h_t^1, h_t^2, ... h_t^n\}$ where $h_t = concat(h_t^2, ... h_t^n)$) as the architecture is general to any number of components.
However, this would result in $length(\mathbf{c})$ number of decoders.
We relax the constraint for dynamic elements by letting $v^k = MLP(h_t)\in \mathbb{R}^{H_1\times H_1\times 32}$ rather than stacking $v^k$ across the spatial dimension.

\textbf{Discriminator:}\enspace
Increasing the capacity of rendering engine alone would not produce better quality images if the capacity of discriminators is not enough such that they can be easily fooled.
We replace the shared encoder of discriminators with a stack of residual blocks shown in Figure \ref{fig:resblock_d}.
Each frame fed through the new encoder results in a $N\times N\times D$ tensor where $N=4$ for VizDoom and $N=5$ for Pacman.
The single frame discriminator is implemented as [Sum(N,N), Linear(1)] where Sum(N,N) sums over the $N\times N$ spatial dimension of the tensor.
The action-conditioned and temporal discriminators use similar architectures as in \ref{sec:supp_discriminator}.

Figure \ref{fig:res_pacman_rollout} shows rollouts on Pacman trained with the higher capacity \Name model.
It clearly shows that the higher capacity model can produce more realistic sharp images.
Consequently, it would be more suitable for future works that involve simulating high fidelity real-world environments.

\begin{figure*}
\begin{center}
\includegraphics[width=\textwidth]{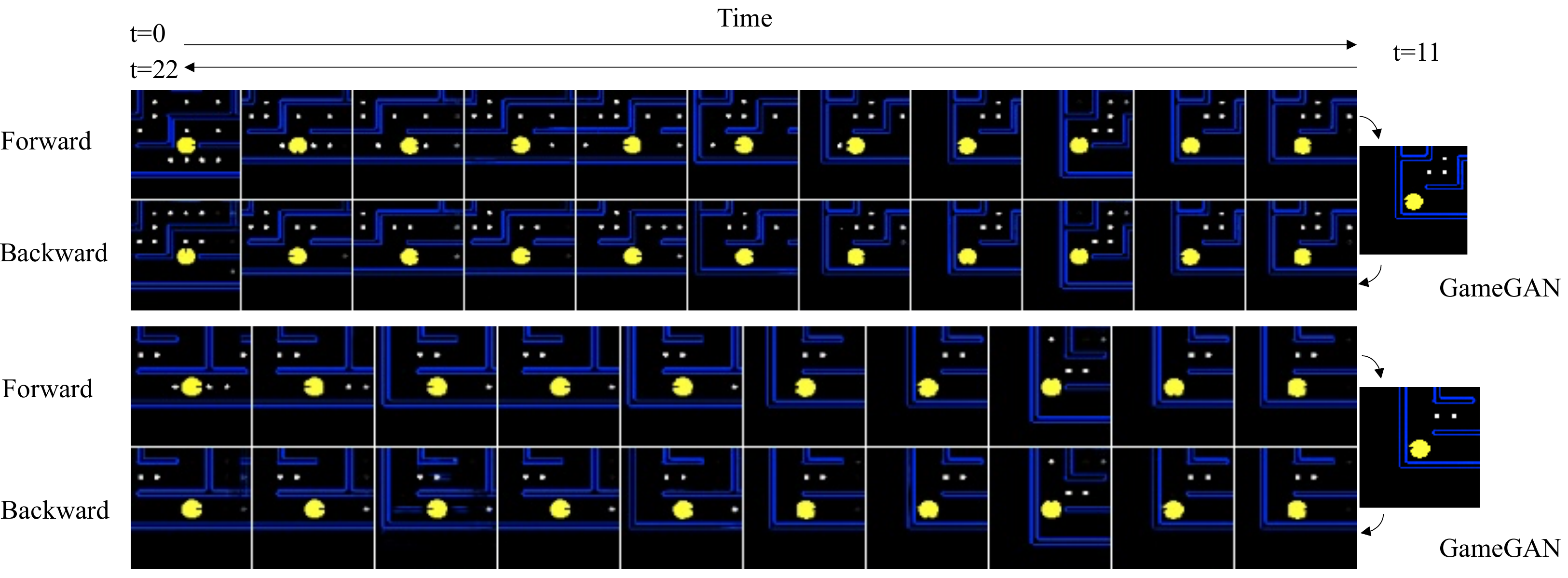}
\end{center}
\caption{Additional Come-back-home task rollouts.
The top row shows the forward path going from the initial position to the goal position.
The bottom row shows the backward path coming back from the goal position to the initial position.
 }
\label{fig:supp_pacman_maze_rollout}
\end{figure*}

\begin{figure*}
\begin{center}
\includegraphics[width=\textwidth]{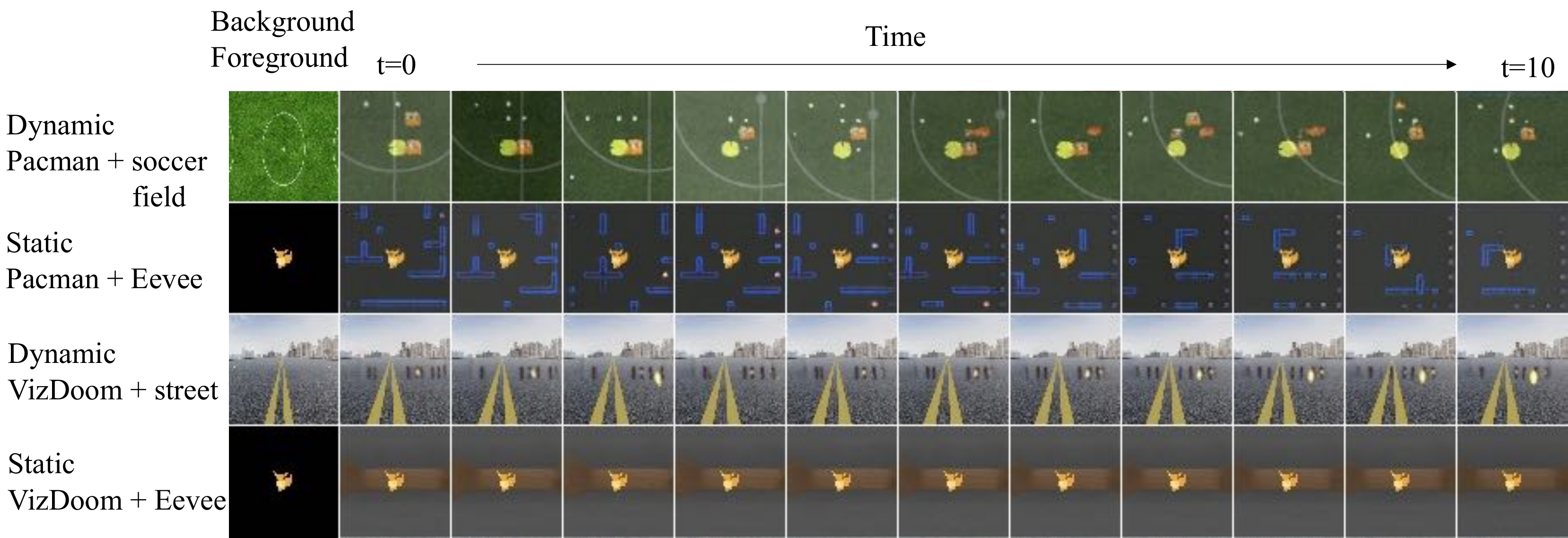}
\end{center}
\caption{Additional rollouts of \Name by swapping background/foreground with random images. }
\label{fig:supp_swap_rollout}
\end{figure*}

\section{Training Scheme}
\label{sec:supp_training}

We employ the standard GAN formulation \cite{goodfellow2014generative} that plays a min-max game between the generator G (in our case, \Name) and the discriminators $D$.
Let $\mathcal{L}_{\mathrm{GAN}}$ be the sum of all GAN losses (we use equal weighting for single frame, action-conditioned, and temporal losses).
Since conditional GAN architectures \cite{mirza2014conditional} are known for learning simplified distributions ignoring the latent code \cite{yang2019diversity, salimans2016improved}, we add information regularization \cite{chen2016infogan} $\mathcal{L}_{\mathrm{Info}}$ that maximizes the mutual information $I(z_t, \phi(x_{t}, x_{t+1}))$ between the latent code $z_t$ and the pair $(x_t, x_{t+1})$.
To help the action-conditioned discriminator, we add a term that minimizes the cross entropy loss $\mathcal{L}_{\mathrm{Action}}$ between $a_t$ and $a^{pred}_t = \psi(x_{t+1}, x_t)$.
Both $\phi$ and $\psi$ are MLP that share layers with the action-conditioned discriminator except for the last layer.
Lastly, we found adding a small L2 reconstruction losses in the image ($\mathcal{L}_{\mathrm{recon}} = \frac{1}{T}\sum_{t=0}^{T} ||x_t - \hat{x}_t ||^2_2 $) and feature spaces ($\mathcal{L}_{\mathrm{feat}} = \frac{1}{T}\sum_{t=0}^{T} ||feat_t - \hat{feat}_t ||^2_2 $) helps stabilize the training.
$x$ and $\hat{x}$ are the real and generated images, and $feat$ and $\hat{feat}$ are the real and generated features from the shared encoder of discriminators, respectively.

\Name optimizes:
\begin{equation}
\mathcal{L} = \mathcal{L}_{\mathrm{GAN}} + \lambda_{\mathrm{A}}\mathcal{L}_{\mathrm{Action}} + \lambda_{\mathrm{I}}\mathcal{L}_{\mathrm{Info}} + \lambda_{\mathrm{r}}\mathcal{L}_{\mathrm{recon}} + \lambda_{\mathrm{f}}\mathcal{L}_{\mathrm{feat}}
\end{equation}

When the memory module and the specialized rendering engine are used, $\lambda_{\mathrm{c}}\mathcal{L}_{\mathrm{cycle}}$ (Section 3.4.2) is added to $\mathcal{L}$ with $\lambda_{\mathrm{c}}=0.05$.
We do not update the rendering engine with gradients from $\mathcal{L}_{\mathrm{cycle}}$, as the purpose of employing $\mathcal{L}_{\mathrm{cycle}}$ is to help train the dynamics engine and the memory module for long-term consistency.
We set $\lambda_{\mathrm{A}} = \lambda_{\mathrm{I}} =1 $ and $ \lambda_{\mathrm{r}}=\lambda_{\mathrm{f}}=0.05$.
The discriminators are updated after each optimization step of \Name.
When traning the discriminators, we add a term $\gamma L_{GP}$ that penalizes the discriminator gradient on the true data distribution \cite{mescheder2018training} with $\gamma=10$.

We use Adam optimizer \cite{kingma2014adam} with a learning rate of 0.0001 for both \Name and discriminators, and use a batch size of 12.
\Name on Pacman and VizDoom environments are trained with a sequence of 18 and 32 frames, respectively.
We employ a warm-up phase where 9 (for Pacman) / 16 (for VizDoom) real frames are fed into the dynamics engine in the first epoch,
and linearly reduce the number of real frames to 1 by the 20-th epoch (the initial frame $x_0$ is always given).






\end{document}